\documentclass{article}
\usepackage{geometry}
\usepackage{booktabs}
\usepackage[dvipsnames,table,xcdraw]{xcolor}

\newgeometry{vmargin={15mm}, hmargin={32mm}}   % set the margins
\usepackage[utf8]{inputenc}

\usepackage{color}
\usepackage{graphicx}% http://ctan.org/pkg/graphicx
\usepackage{subfig}
\usepackage{url}
\usepackage[nopostdot]{glossaries}

\usepackage{textcomp}
\usepackage{algorithm}
\usepackage[noend]{algpseudocode}

\makeglossaries
\setglossarystyle{altlist}

\title{An Energy-Efficient Edge Computing Paradigm for Convolution-based Image Upsampling}
\makeatletter
\renewcommand\@date{{%
  \vspace{-\baselineskip}%
  \large\centering
  \begin{tabular}{@{}c@{}}
    Ian Colbert\textsuperscript{1*} \\
    \normalsize icolbert@eng.ucsd.edu
  \end{tabular}%
  \quad
  \begin{tabular}{@{}c@{}}
    Ken Kreutz-Delgado\textsuperscript{1} \\
    \normalsize kreutz@eng.ucsd.edu 
  \end{tabular}
  \quad
  \begin{tabular}{@{}c@{}}
    Srinjoy Das\textsuperscript{2\footnote{Corresponding authors}} \\
    \normalsize s2das@ucsd.edu
  \end{tabular}

  \bigskip

  \small{\textsuperscript{1}Department of Electrical and Computer Engineering, University of California, San Diego.\par
  \textsuperscript{2}Department of Mathematics, University of California, San Diego.}

  \bigskip

  \today
}}
\makeatother

\newcommand{\Note}{\textcolor{black}}

\begin{document}

\maketitle

\begin{abstract}
A novel energy-efficient edge computing paradigm is proposed for real-time deep learning-based image upsampling applications.
State-of-the-art deep learning solutions for image upsampling are currently trained using either resize or sub-pixel convolution to learn kernels that generate high fidelity images with minimal artifacts.
However, performing inference with these learned convolution kernels requires memory-intensive feature map transformations that dominate time and energy costs in real-time applications. 
To alleviate this pressure on memory bandwidth, we confine the use of resize or sub-pixel convolution to training in the cloud by transforming learned convolution kernels to deconvolution kernels before deploying them for inference as a functionally equivalent deconvolution.
These kernel transformations, intended as a one-time cost when shifting from training to inference, enable a systems designer to use each algorithm in their optimal context by preserving the image fidelity learned when training in the cloud while minimizing data transfer penalties during inference at the edge.
We also explore existing variants of deconvolution inference algorithms and introduce a novel variant for consideration.
We analyze and compare the inference properties of convolution-based upsampling algorithms using a quantitative model of incurred time and energy costs and show that using deconvolution for inference at the edge improves both system latency and energy efficiency when compared to their sub-pixel or resize convolution counterparts.
\end{abstract}

\section{Introduction}

\begin{figure}[t!]
    \centering
    \includegraphics[width=1\linewidth]{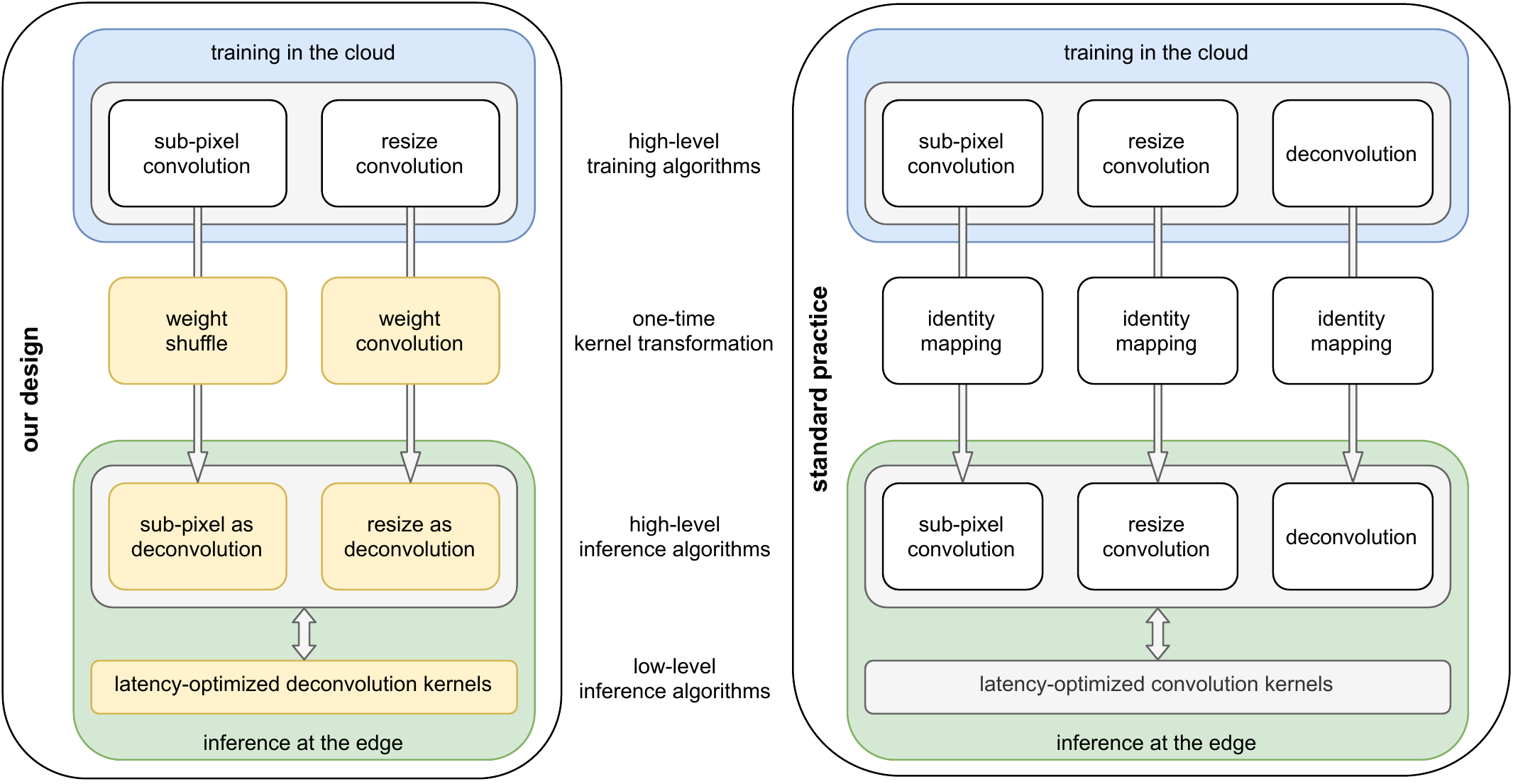}
    \caption{\small{\textbf{A Novel Edge Computing Paradigm for Convolution-based Image Upsampling.}
    In our design paradigm (left), we introduce the blocks highlighted in \textcolor{BurntOrange}{yellow} to use convolution-based image upsampling algorithms in their optimal context, effectively minimizing data transfer penalties when inferencing at the edge while preserving the image fidelity learned when training in the cloud.
    Standard edge computing frameworks (right) deploy pre-trained networks to inference locally on edge devices using the same \Note{high-level} algorithms used in training, \textit{i.e.} an identity mapping from training to inference.}}
    \label{fig:main_idea}
\end{figure}

When building deep learning solutions for latency-sensitive image upsampling problems such as real-time super resolution, systems designers are forced to balance trade-offs between image fidelity and hardware performance.
Models trained using resize convolution~\cite{odena2016deconvolution} learn to upsample images without introducing checkerboard artifacts, but rely on memory-intensive pre-processing to then inefficiently execute compute operations in a higher dimensional space where the cost is greater~\cite{dong2016accelerating,shi2016real}.
Models trained with sub-pixel convolution~\cite{shi2016real} converge faster with less test error when properly initialized~\cite{aitken2017checkerboard}, but require even more memory-intensive post-processing with every inference pass.
Using deconvolution~\cite{zeiler2010deconvolutional}, a model can efficiently generate images without any additional data processing, but training can introduce checkerboard artifacts with gradient updates~\cite{aitken2017checkerboard,odena2016deconvolution}.
% \Note{When considering an end-to-end machine learning pipeline, there is no convolution-based image upsampling algorithm that is superior for both training and inference}
We propose a novel edge computing paradigm that eases the selection between these algorithms by using each in their optimal context.
As depicted in Figure~\ref{fig:main_idea}, our framework confines the use of sub-pixel or resize convolution to training in the cloud, where the cost of their memory-intensive feature map transformations is less severe.
The learned convolution kernels are then transformed to deconvolution kernels, effectively reducing data transfer penalties without sacrificing image fidelity when deployed for energy-efficient inference at the edge as a functionally equivalent deconvolution.

The goal of this paper is to synthesize a collection of previous works into an edge computing design methodology to support real-time convolution-based image upsampling applications\footnote{In this paper, we use \textit{convolution-based upsampling} to denote algorithms relying on either convolution or transposed convolution, commonly known as deconvolution.}.
In such applications, inference at the edge is physically separated from cloud-based training.
This \textit{cloud-to-edge} separation of hardware is currently standard practice and reduces strain on network bandwidth, decreases system latency, and improves overall security~\cite{dhar2019device}.
As shown in Figure~\ref{fig:main_idea}, standard edge computing frameworks deploy pre-trained neural networks to execute locally on edge devices and the high-level algorithms used during training are the same as those used during inference.
% We refer to these high-level algorithms as \textit{front-end} algorithms, as they are often intentionally selected by a systems designer.
\Note{During inference,} these \Note{high-level} algorithms are executed directly as latency-optimized compute kernels often selected by a runtime optimizer without explicit directive from a systems designer.
We refer to these compute kernels as \Note{\textit{low-level algorithms}} so as to not confuse them with learned convolution weights, which are also referred to as kernels.
Our proposed edge computing paradigm enables the use of  \Note{low-level} deconvolution algorithms as inference solutions for real-time image upsampling applications without sacrificing image fidelity.
\Note{Under our paradigm, the high-level algorithms used for training are not the same as those used for inference.
By doing so, we significantly reduce data transfer penalties to improve both latency and energy efficiency during inference at the edge.}
Below, we summarize our contributions\footnote{Find all algorithms described in this paper at \url{https://github.com/icolbert/upsampling}}.

\begin{enumerate}
    \item We enable the physical separation of \Note{high-level} training algorithms from \Note{high-level} inference algorithms by introducing a set of single-use kernel transformations that translate sub-pixel and nearest neighbor resize convolutions to functionally equivalent deconvolutions (Section~\ref{sec:translation-algos}).
    
    \item We provide a comprehensive analysis of existing formulations of deconvolution inference solutions and explore their use as \Note{low-level} algorithms at the edge (Section~\ref{sec:deconvolution}).

    \item We introduce a novel variant of the reverse looping deconvolution algorithm~\cite{zhang2017design} that exposes more opportunities for concurrent execution and improves its adaptability to the limited resource availability of edge platforms (Section~\ref{sec:revd2}).

    \item We analyze and compare the properties of these algorithms under a quantitative model to verify our design paradigm to show that translating to deconvolution for inference at the edge from sub-pixel or resize convolution in the cloud reduces time and energy costs (Section~\ref{sec:experimental-results}).

    \item \Note{We summarize the implications of these experiments and provide recommendations for system designers to support real-time, energy-efficient convolution-based upsampling (Section~\ref{sec:discussion}).}
\end{enumerate}

\section{Convolution-based Image Upsampling Algorithms}
\label{sec:conv_upsampling}

\begin{figure}[t!]
    \centering
    \includegraphics[width=\linewidth]{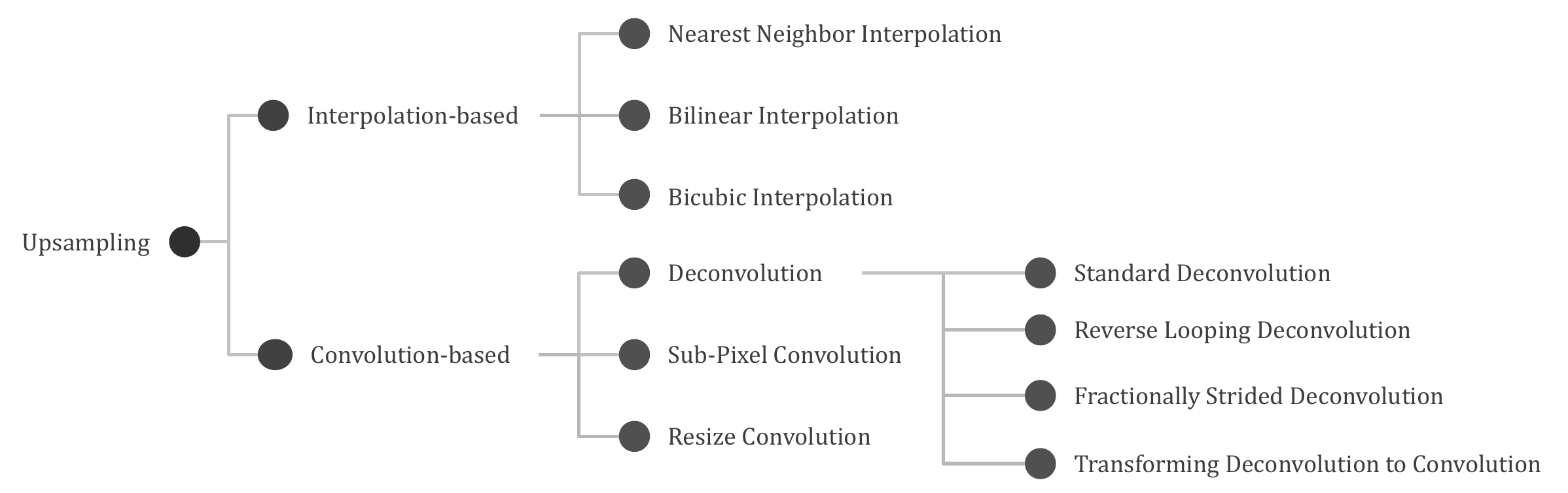}
    \label{fig:downsampling-upsampling}
    \caption{\small{\textbf{Image Upsampling Taxonomy.} State-of-the-art deep learning solutions for image upsampling rely on convolution-based and/or interpolation-based algorithms to increase the resolution of images.}}
\end{figure}

Many solutions to important computer vision and image processing applications require increasing the number of pixels per unit area (or resolution) by inferring values in high dimensional spaces from low dimensional representations - \textit{e.g.} scene segmentation~\cite{long2015fully}, pose estimation~\cite{sun2019deep}, image generation~\cite{goodfellow2014generative}, or super resolution~\cite{shi2016real}.
This process, commonly referred to as upsampling, is a one-to-many mapping to predict, generate, or recover information to increase dimensionality~\cite{shi2016real, wang2020deep}.
In contrast, downsampling is a many-to-one mapping used to encode features and reduce dimensionality~\cite{wang2020deep}.
Deep learning upsampling frameworks typically rely on convolution-based and/or interpolation-based algorithms to increase resolution:
% While the majority of computational convolution primitives focus on optimizing algorithms for downsampling operations typically found in CNNs, there has been little done for upsampling operations
\begin{itemize}
    \item \textbf{Interpolation-based algorithms} infer the value of pixels for which there are no sample points using only local information such as nearest pixel value.
    This class of techniques includes algorithms such as nearest neighbor and bilinear interpolation.
    % In deep neural networks (DNNs), these are typically only used before or after parameterized layers, not in between~\cite{wang2020deep}.
    % As suggested in~\cite{dong2015image} and~\cite{shi2016real}, using non-parametric upsampling techniques before a convolution layer is a special case of a deconvolution layer.

    \item \textbf{Convolution-based upsampling algorithms} also infer the value of pixels for which there are no sample points but predict, generate, or recover high frequency information by learning spatial correlations through training strategies~\cite{shi2016real}.
    This class of techniques includes algorithms such as deconvolution, sub-pixel convolution, and resize convolution, which are commonly used in end-to-end deep learning solutions~\cite{aitken2017checkerboard,dong2015image,dong2016accelerating,shi2016real,odena2016deconvolution,wang2020deep}.
\end{itemize}

\noindent The majority of state-of-the-art deep learning solutions for image upsampling rely on either sub-pixel or resize convolution~\cite{aitken2017checkerboard, dai2019second, dong2015image, odena2016deconvolution, shi2016real, zhang2019residual}.
However, these algorithms require memory-intensive feature map transformations at each inference pass.
In Section~\ref{sec:translation-algos}, we introduce kernel transformations that exploit the functional equivalence of sub-pixel and resize convolution to deconvolution.

% \Rework{It is important to separately consider these operations in training and inference.
% As shown in Section~\ref{sec:translation-algos}, the sub-pixel convolution and resize convolution operations are both functionally equivalent to deconvolution.
% We focus on the analysis of convolution-based upsampling algorithms in the context of energy-efficient edge computing.
% These algorithms can be realized as optimized kernels and implemented as high-performance back-end algorithms~\cite{jorda2019performance,lavin2016fast}.

\subsection{Sub-Pixel Convolution}
\label{sec:sub-pixel-convolution}

\begin{figure}[h!]
    \centering
    \subfloat[$t_0$]{\includegraphics[width=0.13\linewidth]{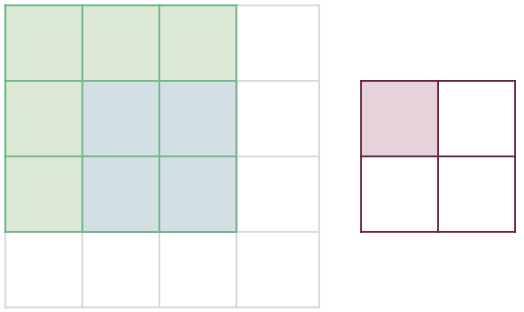}}
    ~~~~~~
    \subfloat[$t_1$]{\includegraphics[width=0.13\linewidth]{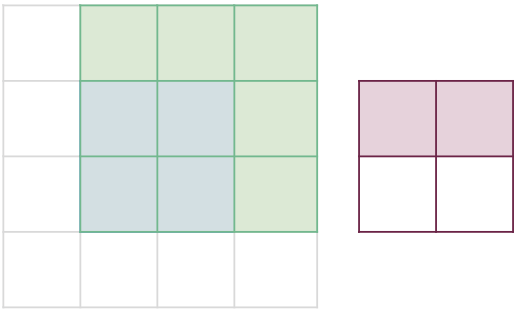}}
    ~~~~~~
    \subfloat[$t_2$]{\includegraphics[width=0.13\linewidth]{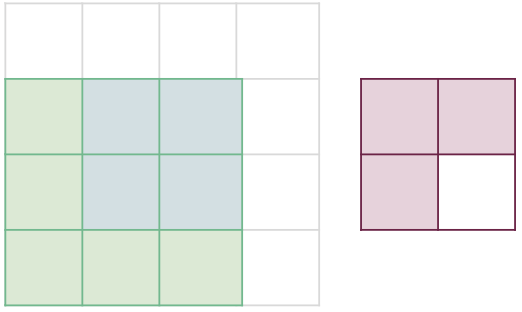}}
    ~~~~~~
    \subfloat[$t_3$]{\includegraphics[width=0.13\linewidth]{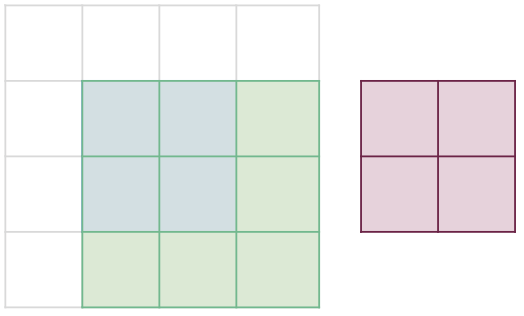}}
    ~~~~~~
    \subfloat[PS]{
    \includegraphics[width=0.13\linewidth]{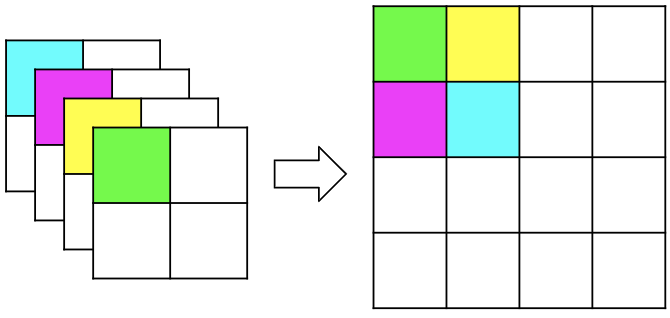}
    \label{subfig:pixel-shuffle}
    }
    \caption{\small{\textbf{Sub-pixel Convolution.} In this example, the $2 \times 2$ input (in \textcolor{NavyBlue}{blue}) is convolved with a $3 \times 3$ kernel (in \textcolor{ForestGreen}{green}) using a stride $S=1$ and padding $P=1$ to create the $2 \times 2$ output (in \textcolor{red}{red}) before upsampling the image by a factor of $2$ using the pixel shuffle (PS). The pixel shuffle is a memory-intensive post-processing operation that introduces significant overhead when used for single-batch inference (Section~\ref{sec:experimental-results}).}}
    \label{fig:subpixel-convolution}
\end{figure}

The sub-pixel convolution was introduced by Shi~\textit{et al.}~\cite{shi2016real} to upsample images using a fully convolutional neural network and is used as the standard method for upsampling images in deep learning solutions~\cite{watson2020deep, wang2020deep}.
As shown in Figure~\ref{fig:subpixel-convolution}, the algorithm is executed as two serialized operations: (1) a same-padded convolution followed by (2) a pixel shuffle.
% Alternatively, it can also be interpreted as a convolution in sub-pixel space and where the transformation between the sub-pixel space and the real-world image is the Pixel Shuffle~\cite{aitken2017checkerboard,shi2016real}.
% As shown in Figure~\ref{fig:subpixel-convolution}, this operation first strides over the output space by convolution before reshaping the resulting feature maps using the pixel shuffle (PS) algorithm depicted in Figure~\ref{subfig:pixel-shuffle}.
However, convolution (see Algorithm~\ref{alg:standard-convolution}) is inherently a downsampling operation.
To upsample by a factor of $r$, the sub-pixel convolution first generates $r^2$ more output channels using a same-padded convolution to then feed the resulting output into the pixel shuffle (see Algorithm~\ref{alg:pixel-shuffle}) to be reshaped.
% As shown in Section~\ref{sec:sub-pixel-convolution}, this is a special case of deconvolution.
Aitken \textit{et al.}~\cite{aitken2017checkerboard} show that, when properly initialized, a network trained using the sub-pixel convolution converges faster with lower image reconstruction error than other convolution-based upsampling algorithms.
However, the sub-pixel convolution requires pixel shuffle post-processing for every inference pass.
As further discussed in Section~\ref{sec:experimental-results}, this memory-intensive feature map transformation severely limits energy efficiency.

% The standard convolution layer as given by Algorithm~\ref{alg:standard-convolution} is a downsampling operation.
% It takes in an input feature map of size $I_H \times I_W \times I_C$ and generates an output feature map of size $O_H \times O_W \times O_C$ using kernel weights of size $K \times K \times I_C \times O_C$.

% This keeps the input and output feature map sizes the same such that $I_H = O_H = H$.

% To upsample by a factor of $r$, the input and output to the convolution operation have dimensions $H\times W\times C$ and $H\times W\times C * r^2$, respectively.
% The output of the convolution layer is then reshaped to $rH\times rW\times C$ using the pixel shuffle.
% Ignoring address calculations of pixel indices, the total number of operations for the sub-pixel convolution to scale an input image by $r$ becomes $r^2 \times K^2 \times C^2 \times H \times W$.

% \begin{algorithm}[h!]
% \caption{Sub-Pixel Convolution}\label{alg:sub-pixel-conv}
% \begin{algorithmic}[1]
% \Procedure{Sub-Pixel Convolution}{}
% \State $h \leftarrow$ Convolution$(x)$
% \State $y \leftarrow$ PixelShuffle$(h)$
% \EndProcedure
% \end{algorithmic}
% \end{algorithm}

\begin{algorithm}[h!]
\caption{Standard Convolution}\label{alg:standard-convolution}
\begin{algorithmic}[1]
\Procedure{Convolution}{}
\For{$o_c = 0$, $o_c{+}{+}$, while $o_c < O_C$}
\For{$o_h = 0$, $o_h{+}{+}$, while $o_h < O_H$}
\For{$o_w = 0$, $o_w{+}{+}$, while $o_w < O_W$}
\For{$i_c = 0$, $i_c{+}{+}$, while $i_c < I_C$}
\For{$k_h = 0$, $k_h{+}{+}$, while $k_h < K$}
\For{$k_w = 0$, $k_w{+}{+}$, while $k_w < K$}
\State $i_h \leftarrow S \times o_h + k_h - P$
\State $i_w \leftarrow S \times o_w + k_w - P$
\State out[$o_c,o_h,o_w$] $\leftarrow$ in[$i_c,i_h,i_w$] $\times$ kernel[$o_c,i_c,k_h,k_w$]
\EndFor
\EndFor
\EndFor
\EndFor
\EndFor
\EndFor
\EndProcedure
\end{algorithmic}
\end{algorithm}

\begin{algorithm}[h!]
\caption{Pixel Shuffle}\label{alg:pixel-shuffle}
\begin{algorithmic}[1]
\Procedure{Pixel Shuffle}{}
\For{$o_c = 0$, $o_c{+}{+}$, while $o_c < O_C$}
\For{$o_w = 0$, $o_w{+}{+}$, while $o_w < O_W$}
\For{$o_h = 0$, $o_h{+}{+}$, while $o_h < O_H$}
\State $i_h \leftarrow \lfloor \frac{o_h}{r} \rfloor$
\State $i_w \leftarrow \lfloor \frac{o_w}{r} \rfloor$
\State $i_c \leftarrow r^2 * o_c +  r ~*~ $\textbf{mod}$(o_h,r) + $\textbf{mod}$(o_w,r)$
\State out[$o_c,o_h,o_w$] $\leftarrow$ in[$i_c,i_h,i_w$]
\EndFor
\EndFor
\EndFor
\EndProcedure
\end{algorithmic}
\end{algorithm}

\subsection{Resize Convolution}
\label{sec:resize-convolution}

\begin{figure}[h!]
    \centering
    \subfloat[NN]{
    \includegraphics[width=0.13\linewidth]{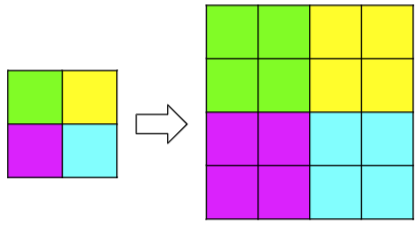}
    \label{subfig:nn-interpolation}
    }
    ~~~~~~
    \subfloat[$t_0$]{\includegraphics[width=0.13\linewidth]{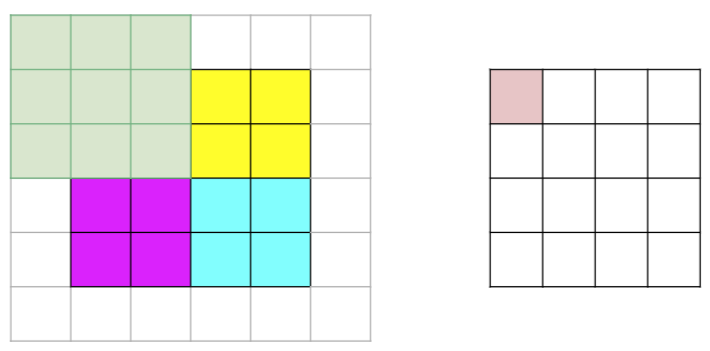}}
    ~~~~~~
    \subfloat[$t_1$]{\includegraphics[width=0.13\linewidth]{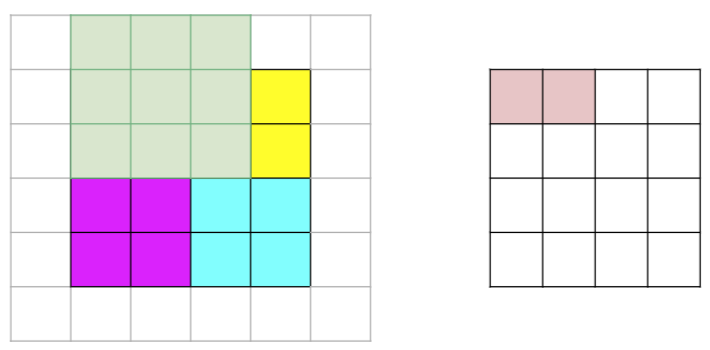}}
    ~~~~~~
    \subfloat[$t_2$]{\includegraphics[width=0.13\linewidth]{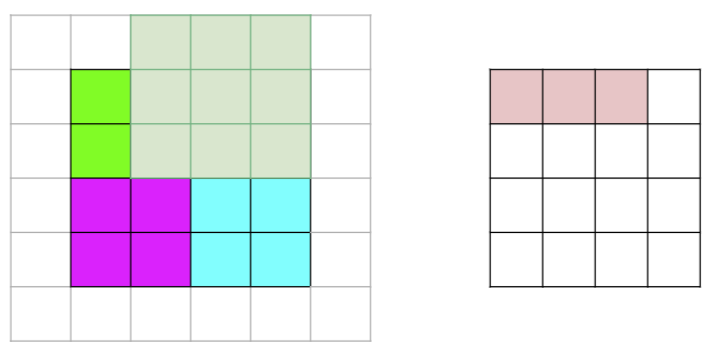}}
    ~~~~~~
    \subfloat[$t_3$]{\includegraphics[width=0.13\linewidth]{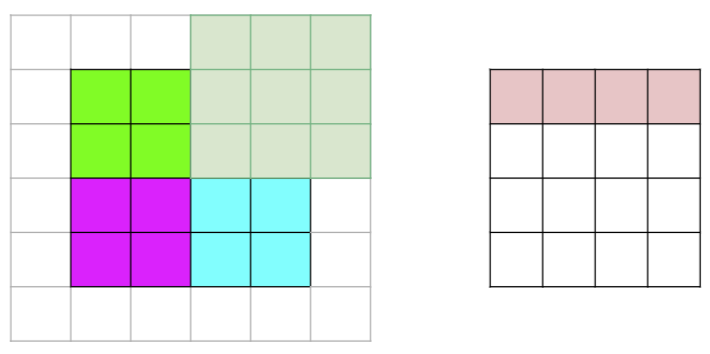}}
    \caption{\small{\textbf{Nearest Neighbor Resize Convolution.} In this example, the $2 \times 2$ input is first upsampled by a factor of 2 using NN interpolation, then convolved with a $3 \times 3$ kernel (in \textcolor{ForestGreen}{green}) in a higher dimensional space to create the $4 \times 4$ output (in \textcolor{red}{red}). NN interpolation is a memory-intensive pre-processing operation that introduces significant overhead when used for single-batch inference (Section~\ref{sec:experimental-results}).}}
    \label{fig:resize-convolution}
\end{figure}

The resize convolution was introduced {by Dong \textit{et al.}~\cite{dong2015image} and popularized} by Odena~\textit{et al.}~\cite{odena2016deconvolution} to address checkerboard artifacts that can arise during training.
As depicted in Figure~\ref{fig:resize-convolution}, the resize convolution is executed as two serialized operations, similar to the sub-pixel convolution.
To upsample by a factor of $r$, the resize convolution first uses (1) an interpolation-based upsampling algorithm before applying (2) a same-padded convolution in the higher resolution space~\cite{odena2016deconvolution}.
Standard implementations for resize convolution rely on nearest neighbor (NN) interpolation as opposed to bilinear or bicubic~\cite{odena2016deconvolution, wang2020deep}.
Similar to sub-pixel convolution's pixel shuffle, the interpolation-based pre-processing is a memory-dominated algorithm required for every inference pass.
Unlike the sub-pixel convolution, the same-padded convolution is executed in a higher dimensional space where the cost of time and energy is much greater~\cite{dong2016accelerating,shi2016real}.
As further discussed in Section~\ref{sec:experimental-results}, this severely limits hardware performance at inference.
% In the context of training, Aitken~\textit{et al.}~\cite{aitken2017checkerboard} show that, by intelligently initializing the kernels of a convolution using a nearest neighbor (NN) strategy, one can effectively circumvent the checkerboard artifact problem while using a sub-pixel convolution.
% By doing so, the NN-initialized sub-pixel convolution converges faster in training and results in superior generative performance than both the standard sub-pixel convolution and NN resize convolution operationns~\cite{aitken2017checkerboard}.
% As shown in Section~\ref{sec:resize-conv-to-deconv}, this is a special case of deconvolution.

% \begin{algorithm}[h!]
% \caption{Resize Convolution}\label{alg:resize-conv}
% \begin{algorithmic}[1]
% \Procedure{Resize Convolution}{}
% \State $h \leftarrow$ Interpolate$(x)$
% \State $y \leftarrow$ Convolution$(h)$
% \EndProcedure
% \end{algorithmic}
% \end{algorithm}

\subsection{Deconvolution}
\label{sec:deconvolution}

\begin{figure}[h!]
    \centering
    \subfloat[$t_0$]{\includegraphics[width=0.18\linewidth]{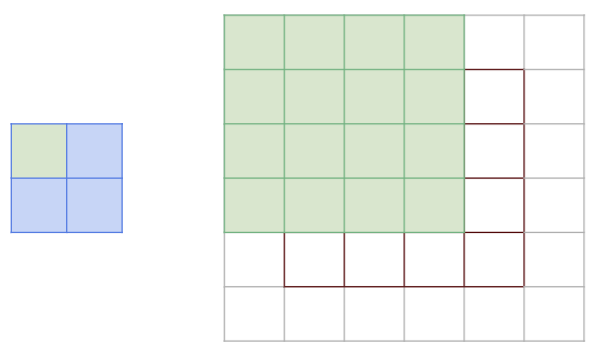}}
    ~~~~~~
    \subfloat[$t_1$]{\includegraphics[width=0.18\linewidth]{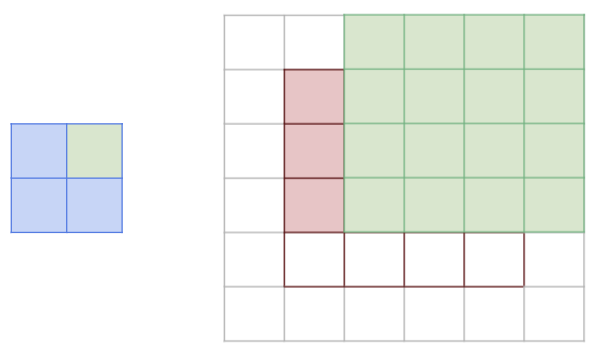}}
    ~~~~~~
    \subfloat[$t_2$]{\includegraphics[width=0.18\linewidth]{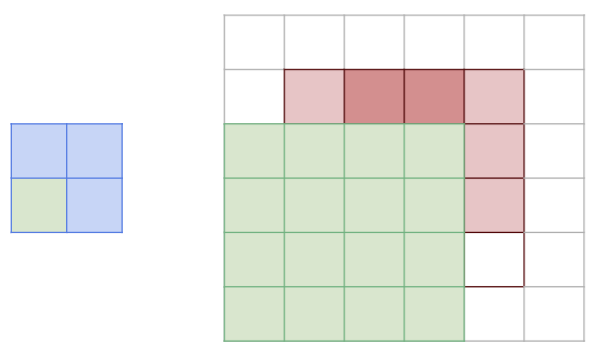}}
    ~~~~~~
    \subfloat[$t_3$]{\includegraphics[width=0.18\linewidth]{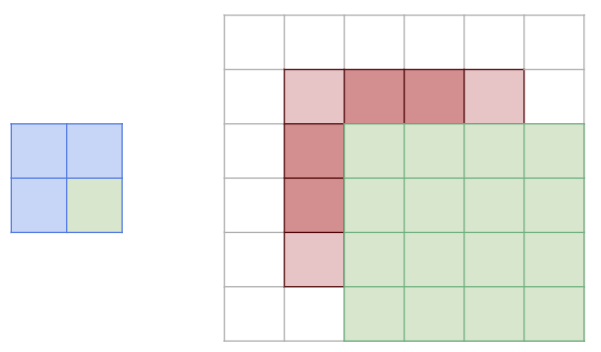}}
    ~~~~~~
    \subfloat[final]{\includegraphics[width=0.18\linewidth]{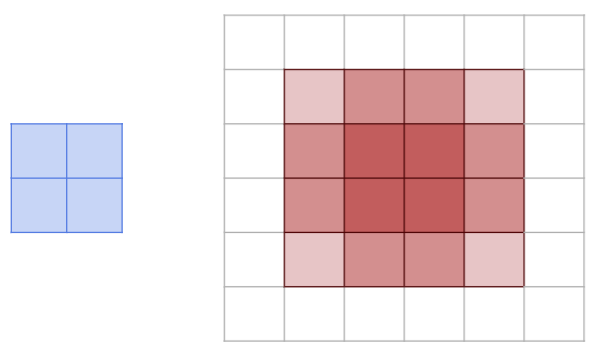}}
    \caption{\small{\textbf{Standard Deconvolution.} The standard deconvolution algorithm strides over the input space, creating overlapping sums in the output space when the stride $S$ is smaller than kernel $K$~\cite{chang2018optimizing,colbert2021competitive,dumoulin2016guide,zhang2017design}. In this example, the $2 \times 2$ input (\textcolor{NavyBlue}{blue}) is deconvolved with a $4 \times 4$ kernel (\textcolor{ForestGreen}{green}) using a stride $S=2$ and a padding $P=1$ to create the $4 \times 4$ output (\textcolor{red}{red}).}}
    \label{fig:deconvolution}
\end{figure}

Deconvolution (also referred to as transpose convolution) is an end-to-end learnable upsampling algorithm that inherently increases the resolution of an input image~\cite{dumoulin2016guide,zeiler2010deconvolutional}.
Deconvolution was first popularized by Zeiler~\textit{et al.}~\cite{zeiler2010deconvolutional} to visualize the latent representations of convolutional neural networks and has since gained popularity in deep learning image upsampling solutions~\cite{wang2020deep}.
Unlike other convolution-based upsampling algorithms, deconvolution upsamples the image directly in one operation.
{A common misconception of the deconvolution operation is that it \textit{requires} inserting zeros to perform fractionally strided convolutions.}
While it is possible to execute deconvolution this way, it greatly increases the input feature map size by adding redundant zero-valued operations, resulting in a much less efficient implementation~\cite{dumoulin2016guide}.
We discuss this formulation further in Section~\ref{sec:strd}.
As shown in Figure~\ref{fig:deconvolution}, the standard deconvolution algorithm given by Algorithm~\ref{alg:stdd} strides over the input space, creating overlapping sums in the output space when the stride $S$ is smaller than kernel $K$~\cite{chang2018optimizing,colbert2021competitive,dumoulin2016guide,zhang2017design}.
{In the context of training, these overlapping sums have been shown to introduce checkerboard artifacts as a result of gradient updates~\cite{odena2016deconvolution}.
In the context of inference, accumulating over these overlapping regions requires storing partial sums when $K > S$ and leads to communication overhead, complex dataflow, and increased resource utilization via on-chip buffering~\cite{chang2018optimizing,colbert2021competitive,liu2018memory,zhang2017design}.
Algorithmic approaches to work around this \textit{overlapping sums} problem are divided into three categories - reverse looping deconvolution, fractionally strided deconvolution, and transforming deconvolution to convolution.}

% In a note following their initial publication proposing the sub-pixel convolution upsampling layer, Shi \textit{et al.}~\cite{shi2016real} suggest that the only difference between the sub-pixel and deconvolution operations are the kernel indices contributing to the final output~\cite{shi2016deconvolution}.
% Given a learned filter, shuffling the indices of learned filters yield identical results - we discuss this more in Section~\ref{sec:algorithm-comparison}.
% In the context of edge computing, this allows for the separation of training from inference and software from hardware without sacrificing image quality.

\begin{algorithm}[h]
\caption{Standard Deconvolution}\label{alg:stdd}
\begin{algorithmic}[1]
\Procedure{Deconvolution}{}
\For{$o_c = 0$, $o_c{+}{+}$, while $o_c < O_C$}
\For{$i_w = 0$, $i_w{+}{+}$, while $i_w < I_W$}
\For{$i_h = 0$, $i_h{+}{+}$, while $i_h < I_H$}
\For{$i_c = 0$, $i_c{+}{+}$, while $i_c < I_C$}
\For{$k_h = 0$, $k_h{+}{+}$, while $k_h < K$}
\For{$k_w = 0$, $k_w{+}{+}$, while $k_w < K$}
\State $o_h \leftarrow S \times i_h + k_h - P$
\State $o_w \leftarrow S \times i_w + k_w - P$
\State out[$o_h,o_w,o_c$] $\leftarrow$ in[$i_h,i_w,i_c$] $\times$ kernel[$i_c,o_c,k_h,k_w$]
\EndFor
\EndFor
\EndFor
\EndFor
\EndFor
\EndFor
\EndProcedure
\end{algorithmic}
\end{algorithm}

\subsubsection{Reverse Looping Deconvolution}

Zhang~\textit{et al.}~\cite{zhang2017design} was the first to propose an efficient deconvolution inference solution for the overlapping sums problem.
Using reverse looping and stride-hole skipping techniques, they traverse the \textit{output} space rather than the input space to expose opportunities for concurrent execution.
The reverse looping deconvolution algorithm (REVD) skips over the output space in $S^2$ independent tiles to be computed concurrently.
The resulting algorithm (see Algorithm~\ref{alg:revd}) relies on expensive modulo arithmetic for address calculations.
Observing that the modulo arithmetic needed to calculate the output pixel address $o_h$ is only dependent on the kernel address $k_h$, Colbert~\textit{et al.}~\cite{colbert2021competitive} minimize its impact by pre-computing and caching these offsets for each value of $k_h$.
Assuming square kernels, the process reduces the number of modulo operations to $2K$, which minimizes resource utilization and on-chip memory as $K$ tends to be small.
In Section~\ref{sec:revd2}, we propose a variant of this algorithm without the use of stride-hole skipping to further expose opportunities for concurrent execution.

% \begin{figure}[h]
%     \centering
%     \subfloat[$t_0$]{\includegraphics[width=0.15\linewidth]{figs/REVD-image-1.png}}
%     ~~~~~~~~
%     \subfloat[$t_1$]{\includegraphics[width=0.15\linewidth]{figs/REVD-image-2.png}}
%     ~~~~~~~~
%     \subfloat[$t_2$]{\includegraphics[width=0.15\linewidth]{figs/REVD-image-3.png}}
%     ~~~~~~~~
%     \subfloat[$t_3$]{\includegraphics[width=0.15\linewidth]{figs/REVD-image-4.png}}
%     ~~~~~~~~
%     \subfloat[final]{\includegraphics[width=0.15\linewidth]{figs/REVD-image-N.png}}
%     \caption{\small{\textbf{Reverse Looping Deconvolution.} To circumvent the overlapping sums problem of the standard deconvolution, the reverse looping deconvolution traverses the output space (\textcolor{red}{red}) using stride-hole skipping to identify dependent inputs (\textcolor{NavyBlue}{blue}) and kernels (\textcolor{ForestGreen}{green}).}}
%     \label{fig:revd_about}
% \end{figure}

\begin{algorithm}
\caption{Reverse Looping Deconvolution (REVD)}\label{alg:revd}
\begin{algorithmic}[1]
\For{$o_c = 0$, $o_c{+}{+}$, while $o_c < O_C$}
\For{$\hat{o}_w = 0$, $\hat{o}_w{+}{=}S$, while $\hat{o}_w < O_W$}
\For{$\hat{o}_h = 0$, $\hat{o}_h{+}{=}S$, while $\hat{o}_h < O_H$}
\For{$k_h = 0$, $k_h{+}{+}$, while $k_h < K$}
\For{$k_w = 0$, $k_w{+}{+}$, while $k_w < K$}
\For{$i_c = 0$, $i_c{+}{+}$, while $i_c < I_C$}
\State $o_h = \hat{o}_h + \textbf{mod}(S - \textbf{mod}(P - k_h, S), S)$
\State $o_w = \hat{o}_w + \textbf{mod}(S - \textbf{mod}(P - k_w, S), S)$
\State $i_h = (o_h + P - k_h)/ S$
\State $i_w = (o_w + P - k_w)/ S$
\State y[$o_h,o_w,o_c$] $\leftarrow$ x[$i_h,i_w,i_c$] $\times$ w[$k_h,k_w,i_c,o_c$]
\EndFor
\EndFor
\EndFor
\EndFor
\EndFor
\EndFor
\end{algorithmic}
\end{algorithm}

\subsubsection{Fractionally Strided Deconvolution}\label{sec:strd}

\begin{figure}[h!]
    \centering
    \subfloat[$t_0$]{\includegraphics[width=0.19\linewidth]{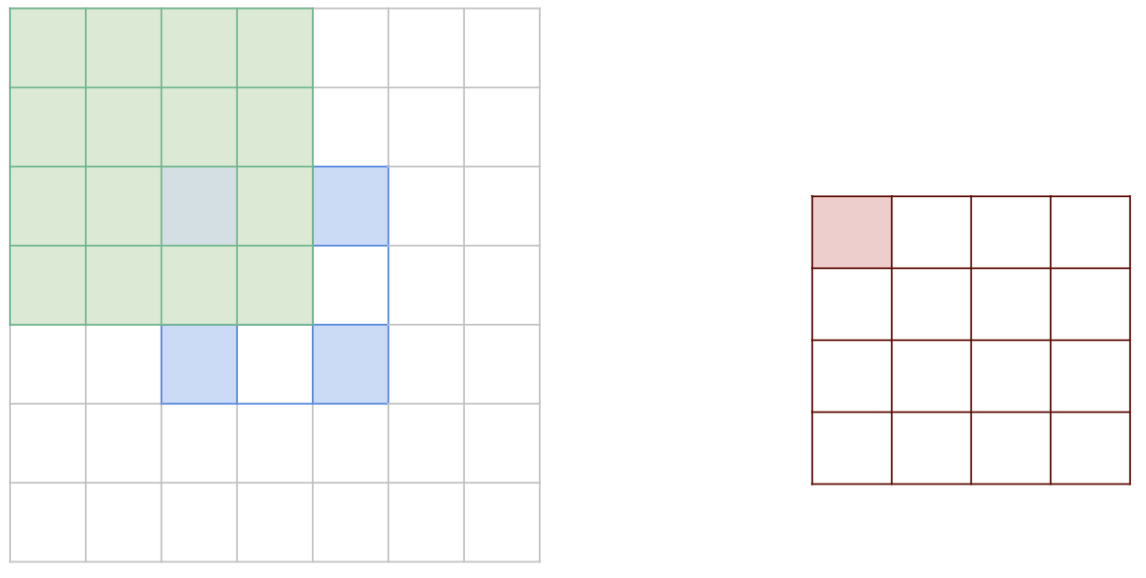}}
    ~~~
    \subfloat[$t_1$]{\includegraphics[width=0.19\linewidth]{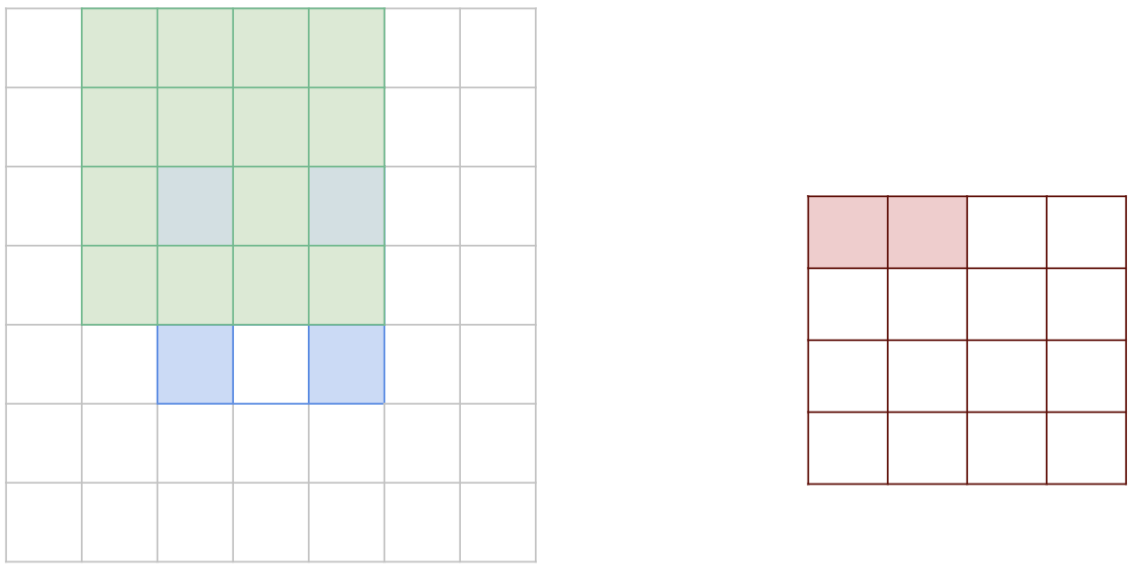}}
    ~~~
    \subfloat[$t_2$]{\includegraphics[width=0.19\linewidth]{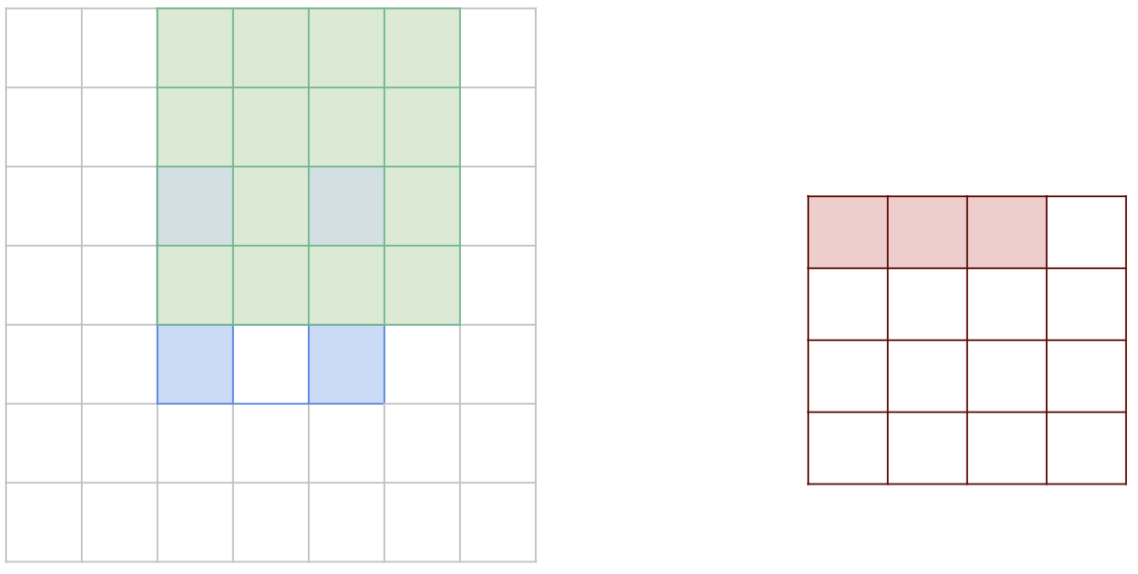}}
    ~~~
    \subfloat[$t_3$]{\includegraphics[width=0.19\linewidth]{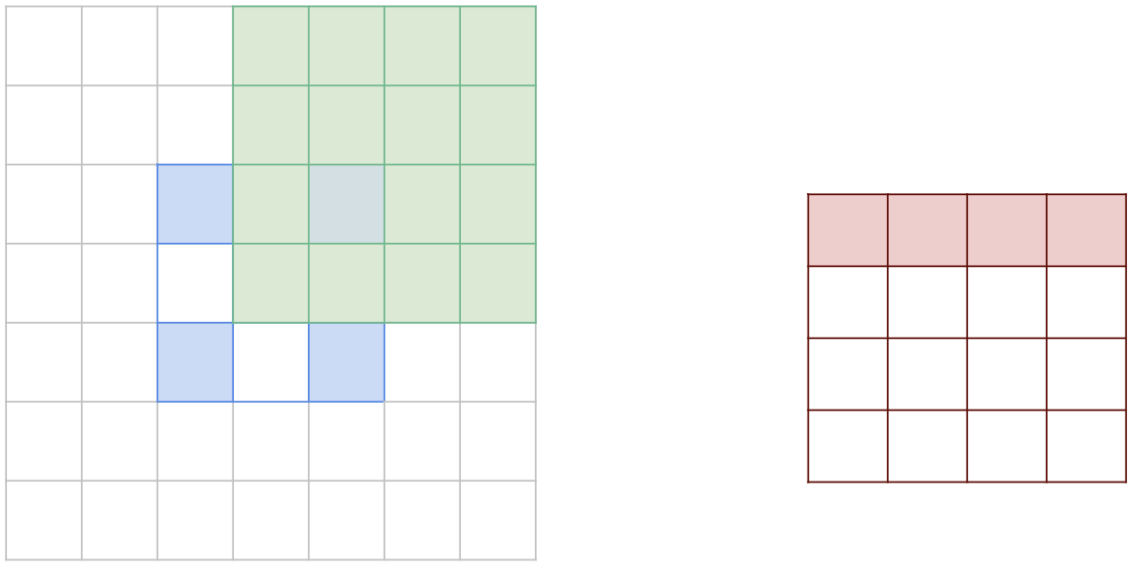}}
    ~~~
    \subfloat[final]{\includegraphics[width=0.19\linewidth]{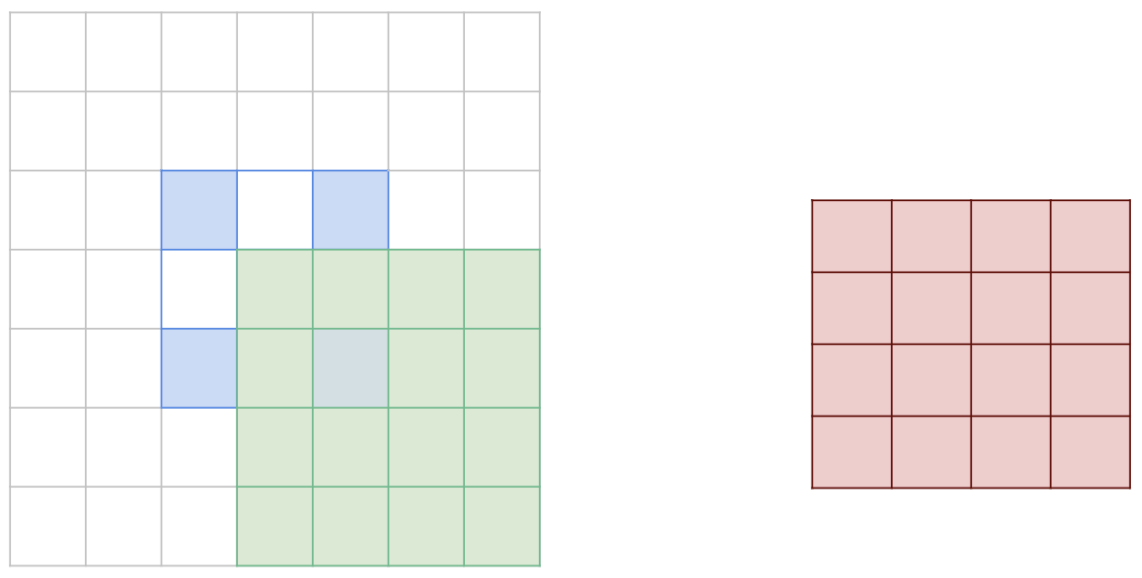}}
    \caption{\small{\textbf{Fractionally Strided Deconvolution.} For the fractionally strided deconvolution to be equivalent to the example in Figure~\ref{fig:deconvolution}, $S - 1$ zeros are first inserted in between the input pixels. Then, the transpose of the $4 \times 4$ deconvolution kernels (\textcolor{ForestGreen}{green}) are convolved over the  $3 \times 3$ input (\textcolor{NavyBlue}{blue}) with a stride of $S=1$ and padding $P = 2$ to create the $4 \times 4$ output (\textcolor{red}{red}).}}
    \label{fig:strd_about}
\end{figure}

% ~\cite{wang2019towards,yazdanbakhsh2018flexigan,yazdanbakhsh2018ganax}
The fractionally strided deconvolution (STRD) can be implemented using unmodified convolution accelerators and is most commonly used by machine learning frameworks such as PyTorch~\cite{NEURIPS2019_9015} and TensorFlow~\cite{tensorflow2015-whitepaper}.
It avoids the overlapping sum problem by padding each input pixel by $S$ zeros before executing a standard convolution using transposed kernels~\cite{dumoulin2016guide}.
As such, it can be viewed as two serialized operations: (1) a zero-insertion feature map transformation followed by (2) a same-padded transposed convolution.
{As shown in Figure~\ref{fig:strd_about}}, to upsample by a factor of $r$, the fractionally strided convolution first inserts $H - 1$ rows and $W - 1$ columns of $S - 1$ zeros into the input feature map~\cite{dumoulin2016guide} before transposing the deconvolution kernels to execute a same-padded convolution.
While this works around the overlapping sum problem, it introduces massive redundancies as the input feature map grows~\cite{dumoulin2016guide}.
Figure~\ref{fig:sparsity_requirements}a shows how the percentage of zero-valued input pixels increases with upsamling factor $r$.

% \begin{figure}
%     \centering
%     \includegraphics[width=0.5\linewidth]{figs/STDR-feature-sparsity.png}
%     \caption{\small{\textbf{Fractionally Strided Deconvolution Zero-Insertion Requirements.} Implementing deconvolution as a fractionally strided deconvolution, as is common practice, greatly increases the input feature map size by adding redundant zero-valued operations, resulting in a much less efficient implementation~\cite{dumoulin2016guide}. Assuming a square input feature map, upsampling by a factor of 2 requires 75\% of input pixels to be zero.}}
%     \label{fig:strd_sparsity_requirements}
% \end{figure}

% \begin{algorithm}[h!]
% \caption{Fractionally Strided Deconvolution}\label{alg:strd}
% \begin{algorithmic}[1]
% \Procedure{Fractionally Strided Deconvolution}{}
% \State $x \leftarrow$ ZeroInsertion$(x)$
% \State $w \leftarrow$ TransposeKernels$(w)$
% \State $y \leftarrow$ Convolution$(x, w)$
% \EndProcedure
% \end{algorithmic}
% \end{algorithm}

\subsubsection{Transforming Deconvolution to Convolution}

\begin{figure}
    \centering
    % \subfloat[STRD]{\includegraphics[width=0.45\linewidth]{figs/STDR-feature-sparsity.png}}
    % \subfloat[TDC]{\includegraphics[width=0.45\linewidth]{figs/TDC-kernel-sparsity.png}}
    \includegraphics[width=0.46\linewidth]{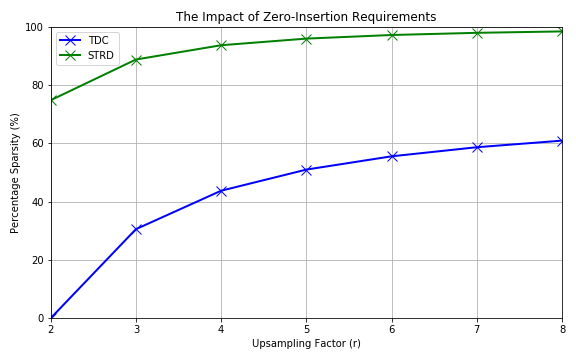}
    \caption{
    \small{
    \textbf{The Impact of Zero-Insertion Requirements for STRD and TDC.}
    As shown in \textbf{\textcolor{ForestGreen}{green}}, implementing deconvolution as a fractionally strided deconvolution (STRD), as is common practice, greatly increases the input feature map size by adding redundant zero-valued operations~\cite{dumoulin2016guide}. Assuming a square input feature map, even upsampling by a factor of 2 requires 75\% of input pixels to be zero.
    As shown in \textbf{\textcolor{blue}{blue}}, transforming deconvolution to a convolution (TDC) requires padding the transformed kernel tiles by $P_K$ when the kernel size $K$ is not evenly divisible by the stride $S$. Assuming a square input feature map and $P_K = 2$, the percentage of zeros increases with upsampling factor $r$.
    }
    }
    \label{fig:sparsity_requirements}
\end{figure}

Chang \textit{et al.}~\cite{chang2018optimizing} avoid the overlapping sum problem by transforming deconvolution into $S^2$ tiled convolutions to compute each region of the output feature map independently.
They refer to this formulation as transforming deconvolution to convolution (TDC).
To split a deconvolution into $S^2$ convolutions, the algorithm first uses the transformation given by Algorithm~\ref{alg:tdc-weights} to split the deconvolution kernels into $S^2$ tiles of size $K_T \times K_T$, where $K_T = \lceil \frac{K}{S} \rceil$.
When $K$ is not evenly divisible by $S$, this transformation requires padding each kernel tile by $P_K$, where $P_K = (S \times K_T) - K$.
Figure~\ref{fig:sparsity_requirements}b shows how the percentage of zero-valued kernels increases with upsampling factor $r$.
Given the transformed kernels, the output pixels of each of the $S^2$ tiles can then be computed concurrently using a same-padded convolution.
However, similar to the sub-pixel convolution, the transformation process requires expensive post-processing to stitch the resulting tiles back together~\cite{chang2018energy,chang2018optimizing,xu2020accelerating}.
Xu \textit{et al.}~\cite{xu2020accelerating} propose a variant that stitches the output tiles during computation, which reduces the cost of data transfers by integrating post-processing arithmetic into the logic of the base algorithm.
In our algorithm comparisons, we focus on this variant of TDC.

\begin{algorithm}
\caption{Transforming Deconvolution to Convolution: Kernel Transformation}\label{alg:tdc-weights}
\begin{algorithmic}[1]
\Procedure{Kernel Transformation}{}
\For{$o_c = 0$, $o_c{+}{+}$, while $o_c < O_C$}
\For{$i_c = 0$, $i_c{+}{+}$, while $i_c < I_C$}
\For{$k_h = 0$, $k_h{+}{+}$, while $k_h < K_H + P_K$}
\For{$k_w = 0$, $k_w{+}{+}$, while $k_w < K_W + P_K$}
\State $n = S * \textbf{mod}(k_h, S) + \textbf{mod}(k_w, S)$
\State $\hat{k}_h = K_T - \lceil \frac{k_h + 1}{S} \rceil$
\State $\hat{k}_w = K_T - \lceil \frac{k_w + 1}{S} \rceil$
\State $w_T[o_c, i_c, n, \hat{k}_h, \hat{k}_w] = w[i_c, o_c, k_h, k_w]$
\EndFor
\EndFor
\EndFor
\EndFor
\EndProcedure
\end{algorithmic}
\end{algorithm}

% \begin{algorithm}
% \caption{Transforming Deconvolution to Convolution: Pixel Value Calculation}\label{alg:tdc}
% \begin{algorithmic}[1]
% \Procedure{Compute}{}
% \For{$o_c = 0$, $o_c{+}{+}$, while $o_c < O_C$}
% \For{$n = 0$, $n{+}{+}$, while $n < S^2$}
% \For{$i_c = 0$, $i_c{+}{+}$, while $i_c < I_C$}
% \For{$\hat{o}_h = 0$, $\hat{o}_h{+}{+}$, while $\hat{o}_h < \lfloor \frac{O_H}{S} \rfloor$}
% \For{$\hat{o}_w = 0$, $\hat{o}_w{+}{+}$, while $\hat{o}_w < \lfloor \frac{O_W}{S} \rfloor$}
% \For{$k_h = 0$, $k_h{+}{+}$, while $k_h < K_T$}
% \For{$k_w = 0$, $k_w{+}{+}$, while $k_w < K_T$}
% \State $o_h = \hat{o}_h * S + \lfloor \frac{n}{S} \rfloor$
% \State $o_w = \hat{o}_w * S + \textbf{mod}(n, S)$
% \State $i_h = \hat{o}_h + k_h$
% \State $i_w = \hat{o}_w + k_w$
% \State y[$o_h,o_w,o_c$] $\leftarrow$ x[$i_h,i_w,i_c$] $\times w_T$[$k_h,k_w,n,i_c,o_c$]
% \EndFor
% \EndFor
% \EndFor
% \EndFor
% \EndFor
% \EndFor
% \EndFor
% \end{algorithmic}
% \end{algorithm}

\section{Improving Reverse Looping Deconvolution}
\label{sec:revd2}

% When using standard convolution or deconvolution arithmetic, each output pixel is data-independent.
% As shown in Algorithm~\ref{alg:standard-convolution}, standard convolution arithmetic traverses the output space and exposes opportunities for extreme data-level parallelism (DLP) both spatially and temporally.
% Assuming sufficient hardware resources, each output pixel can be executed concurrently and pipelined to complete a full forward pass in $I_C \times K^2 + D_p$ cycles, where $D_p$ is the pipeline depth.
% In Section~\ref{sec:algorithm-comparison}, we address how this efficiency is constrained for the sub-pixel convolution by the Pixel Shuffle operator.

% https://docs.oracle.com/cd/E19957-01/805-4940/z400090e5229/index.html
When the memory requirements of a deep learning model exceed the resources available on an edge device, the inference pass is typically divided into smaller workloads using tiling~\cite{ma2017optimizing,zhang2015optimizing}.
These tiled workloads are data-independent if they each write to distinct memory locations without overlap.
Algorithms that can be divided into smaller tiles have higher degrees of parallelism as each data-independent workloads can be executed concurrently across SIMD lanes on multi-threaded hardware.
When data-independent workloads are evenly balanced, tiling optimizations can increase hardware utilization and lead to improved energy efficiency.
% In this paradigm, weights and activations are first read from off-chip memory to then be used to compute each output tile which is written back to off-chip memory.
% To analyze the impact of data-independence on edge devices, we consider load balance - the even distribution of work among all processes in a parallel system.
On the other hand, the presence of imbalanced workloads can force all live processes to wait for an overloaded lane to finish before synchronization~\cite{pearce2012quantifying}.
Algorithms with higher degrees of parallelism expose more opportunities for efficient concurrent execution as workloads are more easily balanced across SIMD lanes.

To understand the impact of data-independence and load balance on inference acceleration, consider a processor with 16 SIMD lanes designed to accelerate a deconvolution workload.
When used to upsample a $14\times14$ image by a factor of $2$ using a stride of $2$ such that $r=2$, $S=2$, and $O_H \times O_W = 28 \times 28$, a designer could use a tile size of $7$ to divide the total work into 16 data-independent workloads to be dispatched across each lane and achieve 100\% load balance.
% The penalty for imbalanced workloads grows with the number of concurrent processes~\cite{pearce2012quantifying}.
However, both the reverse looping deconvolution (REVD) and transforming deconvolution to convolution (TDC) algorithms traverse the output space in $S^2$ tiles~\cite{chang2018optimizing, zhang2017design}.
As a consequence, functional correctness breaks down when the tiling along the output space is not perfectly divisible by the stride $S$.
% {This constrains the degree to which data-independence can be exploited as}
% As $r$ increases, the lower-bound on the tile size increases by $r^2$.
% To analyze the impact of data-independence on edge devices, we consider load balance - the even distribution of work among all processes in a parallel system.
% The presence of imbalanced workloads can force all live processes to wait for an overloaded process to finish before synchronization.
A tiling factor of 7 is not perfectly divisible by the stride of 2 so a designer is left with options 6 and 8.
Using a tiling factor of 8 requires zero-padding each workload, effectively increasing the data movement by 30\%\footnote{Moving 16 workloads of $8\times8$ pixels is 1.3x that of moving 16 workloads of $7\times7$.}.
This 1.3x increase is detrimental to the system's energy efficiency as energy consumption is dominated by data movement~\cite{horowitz20141}.
Using a tiling factor of 6 requires multiplexing through the 16 SIMD lanes twice, reducing hardware utilization to 78\%, increasing system latency, and introducing imbalanced workloads that would need to wait for each lane to finish\footnote{Executing 16 workloads of $6\times6$ pixels only covers ~73\% of the total compute work of a $28\times28$ image. To complete the final 27\%, the hardware would need to multiplex through the SIMD lanes a second time with 6 workloads of $6\times6$ pixels. In this pass, only 36\% of the hardware is utilized. Over both passes, only 78\% of the hardware is being used.}.
% 28x28 real work, (4x8x4x8) total work
% 25 work items, 32 dispatches in SIMD
% To be able to separate each output pixel into data-independent workloads would require that $S = 1$.
% These conditions are only satisfied with an upscaling factor of 1 where $P=S=r=1$, as shown in Section~\ref{sec:converting-subpixel-to-deconv}.
% Recall that, to upsample by a factor of $r$, the deconvolution equivalent to a sub-pixel convolution requires a kernel size of $r*K$.
% Once again assuming sufficient hardware resources to fully execute each data-independent workload concurrently, each output pixel would require $I_C \times r^2 \times K^2 + D_p$ cycles to complete a full forward pass, increasing the cycle count lower-bound by a factor of $1 + r^2$.
% Even a scaling factor of $r=2$ raises this lower bound by $40\%$ (1.4x).
To both circumvent the overlapping sums problem and fully exploit data-independent concurrent execution, a deconvolution algorithm needs to traverse the output space without the use of stride-hole skipping.

We propose an improved reverse looping deconvolution algorithm, which we refer to as REVD2.
% The reverse looping deconvolution arithmetic of REVD traverses the weight space sequentially and uses the kernel index $k_h$ to compute the output index $o_h$ using stride-hole skipping along the output space.
%As shown in Algorithm~\ref{alg:revd2}, REVD2 uses stride-hole skipping along the \textit{weight} space rather than the output space to fully expose the data-independence of output pixels for more effective load balancing while avoiding the overlapping sums problem.
As shown in Algorithm~\ref{alg:revd2}, REVD2 uses stride-hole skipping along the \textit{weight} space rather than the output space to both avoid the overlapping sums problem and fully expose the data-independence of output pixels for more effective load balancing. 
Unlike TDC or REVD, REVD2 supports a tile size of 7 in the example described above.
It also reduces the cost of modulo arithmetic.
We can fully remove any dependence of REVD2 on modulo arithmetic by leveraging the data-independence of each output pixel.
When dispatching each tiled workload for concurrent execution, $\textbf{mod}(o_h + P, S)$ can be replaced by a simple counter $j$ initialized to $P$.
When $j \geq S$, it can be reset to zero\footnote{We have implemented this at \url{https://github.com/icolbert/upsampling}}. 
% % This can lead to significant overhead when realized in embedded hardware as it is both a long-latency operation and greatly affects cache locality~\cite{colbert2021competitive,tu2019accelerating}.
% Assuming sufficient hardware resources, each output pixel would now require $I_C \times r^2 \times K^2 / S^2 + D_p$ cycles when using stride-hole skipping along the weight space.
% When applied to image upscaling, $r = S$ so the lower-bound cycle estimate to complete a forward pass becomes $I_C \times K^2 + D_p$.

\begin{algorithm}
\caption{Improving Reverse Looping Deconvolution (REVD2)}\label{alg:revd2}
\begin{algorithmic}[1]
\For{$o_c = 0, o_c{+}{+}$, while $o_c < O_C$}
\For{$o_h = 0, o_h{+}{+}$, while $o_h < O_H$}
\For{$o_w = 0, o_w{+}{+}$, while $o_w < O_W$}
\For{$i_c = 0, i_c{+}{+}$, while $i_c < I_C$}
\For{$\hat{k}_h = 0, \hat{k}_h{+=}{S}$, while $\hat{k}_h < K$}
\For{$\hat{k}_w = 0, \hat{k}_w{+=}{S}$, while $\hat{k}_w < K$}
\State $k_h = \hat{k}_h + \textbf{mod}(o_h + P, S)$
\State $k_w = \hat{k}_w + \textbf{mod}(o_w + P, S)$
\State $i_h = (o_h + P - k_h) / S$
\State $i_w = (o_w + P - k_w) / S$
\State y[$o_h,o_w,o_c$] $\leftarrow$ x[$i_h,i_w,i_c$] $\times$ w[$k_h,k_w,i_c,o_c$]
\EndFor
\EndFor
\EndFor
\EndFor
\EndFor
\EndFor
\end{algorithmic}
\end{algorithm}

% \begin{algorithm}
% \caption{REVD2 without Modulo Arithmetic}
% \label{alg:revdv2-no-mods}
% \begin{algorithmic}[1]
% \State counter$_{k_h} = $ OffsetCounter$(P,S)$
% \State counter$_{k_w} = $ OffsetCounter$(P,S)$
% \State $f_S \leftarrow 1 / S$
% \For{$o_c = 0$, $o_c{+}{+}$, while $o_c < O_C$}
% \For{$o_h = 0$, $o_h{+}{+}$, while $o_h < O_H$}
% \State $f_h = $counter$_{k_h}$.next()
% \For{$o_w = 0$, $o_w{+}{+}$, while $o_w < O_W$}
% \State $f_w = $counter$_{k_w}$.next()
% \For{$i_c = 0$, $i_c{+}{+}$, while $i_c < I_C$}
% \For{$\hat{k}_h = 0$, $\hat{k}_h{+=}{S}$, while $\hat{k}_h < K$}
% \For{$\hat{k}_w = 0$, $\hat{k}_w{+=}{S}$, while $\hat{k}_w < K$}
% % \State $k_h = \hat{k}_h + \textbf{mod}(o_h + P, S)$
% % \State $k_w = \hat{k}_w + \textbf{mod}(o_w + P, S)$
% \State $k_h = \hat{k}_h + f_h$
% \State $k_w = \hat{k}_w + f_h$
% \State $i_h = (o_h + P - k_h) * f_S$
% \State $i_w = (o_w + P - k_w) * f_S$
% \State y[$o_h,o_w,o_c$] $\leftarrow$ x[$i_h,i_w,i_c$] $\times$ w[$k_h,k_w,i_c,o_c$]
% \EndFor
% \EndFor
% \EndFor
% \EndFor
% \EndFor
% \EndFor
% \end{algorithmic}
% \end{algorithm}

% \begin{algorithm}
% \caption{OffsetCounter}
% \label{alg:offset-counter}
% \hspace*{\algorithmicindent} \textbf{Inputs:} $v_0 \leftarrow$ initial offset value, $C \leftarrow$ maximum offset value \\
% \hspace*{\algorithmicindent} \textbf{Output:} $v_t \leftarrow$ current offset value
% \begin{algorithmic}[1]
% \Procedure{Next}{}
% \State $v_t = c_v$
% \State $c_v \leftarrow v_t + 1$
% \If{$c_v >= C$}
% \State $c_v \leftarrow 0$
% \EndIf
% \Return $v_t$
% \EndProcedure
% \end{algorithmic}
% \end{algorithm}

\section{Kernel Transformations for Deconvolution Inference}
\label{sec:translation-algos}

State-of-the-art deep learning solutions for image upsampling are currently trained using either sub-pixel or resize convolution to learn kernels that generate high fidelity images with minimal artifacts~\cite{aitken2017checkerboard, dai2019second, dong2015image, odena2016deconvolution, shi2016real, zhang2019residual}.
However, convolution is inherently a downsampling algorithm.
As discussed in Section~\ref{sec:conv_upsampling}, inferencing with these learned convolution kernels requires memory-intensive feature map transformations to upsample an image.
Alternatively, deconvolution is inherently an upsampling algorithm.
As discussed in Section~\ref{sec:deconvolution}, the standard deconvolution increases the resolution of an image without reliance on extraneous feature map transformations.
To preserve the image fidelity learned through training while avoiding the data transfer penalties during inference, we introduce two novel kernel transformation algorithms that exploit the functional equivalence of deconvolution to these two state-of-the-art convolution-based upsampling algorithms.
\Note{As opposed to the feature map transformations required for each sub-pixel or resize convolution inference pass,} these kernel transformations are intended as a one-time sunk cost in software \textit{before} deploying the trained model for inference and, once deployed, can be executed using any of the deconvolution formulations described in Section~\ref{sec:deconvolution}\footnote{The TDC formulation of deconvolution does not undo these kernel transformations. It splits the resulting deconvolution into $S^2$ tiled convolutions to be executed concurrently and stitched back together. While the algorithm is functionally equivalent to deconvolution, it may require zero-padding the kernels as described in Section~\ref{sec:deconvolution}.}.

Given that functions $f$ and $g$ are respectively parameterized by $\theta$ and $\beta$, then $g_\beta$ is functionally equivalent to $f_\theta$ if $f_\theta(x) = g_\beta(x)$ for all valid $x$.
As described in Section~\ref{sec:conv_upsampling}, both the sub-pixel and resize convolution use same-padded convolutions, which use a stride of 1.
As shown below, this restricts the valid convolution kernels sizes to be odd as $K = 2P + 1$.
When transforming learned convolution kernels to deconvolution kernels for inference at the edge, the functional equivalence holds for these valid kernels sizes, \textit{e.g.} 3, 5, 7, 9.
\vspace{-0.1cm}
$$
O_H = (I_H - K + 2P) / S + 1
$$
$$
H = (H - K + 2P) / 1 + 1
$$
\begin{equation}\label{eq:conv_kernel_size}
K = 2P + 1   
\end{equation}

\vspace{-0.5cm}
\subsection{Sub-Pixel Convolution to Deconvolution}
\label{sec:converting-subpixel-to-deconv}

To upsample by a factor of $r$, the sub-pixel convolution generates $r^2$ more output channels before applying the pixel shuffle algorithm over the output space.
As shown in Figure~\ref{fig:subpixel-convolution}, this results in a unique pattern of $r^2$ output pixels each originating from independent channels.
To replicate this pattern, a functionally equivalent deconvolution uses a stride $S = r$ with a kernel size $K^D = rK$.
% which becomes an important design decision to consider when training an upsampling network using the sub-pixel convolution and running inference on a hardware accelerator as a deconvolution, as explained below.
The deconvolution padding $P^D$ is calculated below where $K$ and $P$ are the convolution kernel size and padding given in Eq.~\ref{eq:conv_kernel_size},  $I_H$ is the input height, $K^D$ is the deconvolution kernel size, $S$ is the stride, and $O_H$ is the output height~\cite{dumoulin2016guide}.

$$
O_H = S \times ( I_H - 1 ) + K^D - 2P^D
$$
$$
rH = r(H - 1) + rK - 2P^D
$$
$$
2P^D = r(K - 1)
$$
$$
P^D = rP
$$

\begin{figure}[h]
    \centering
    \includegraphics[width=0.45\linewidth]{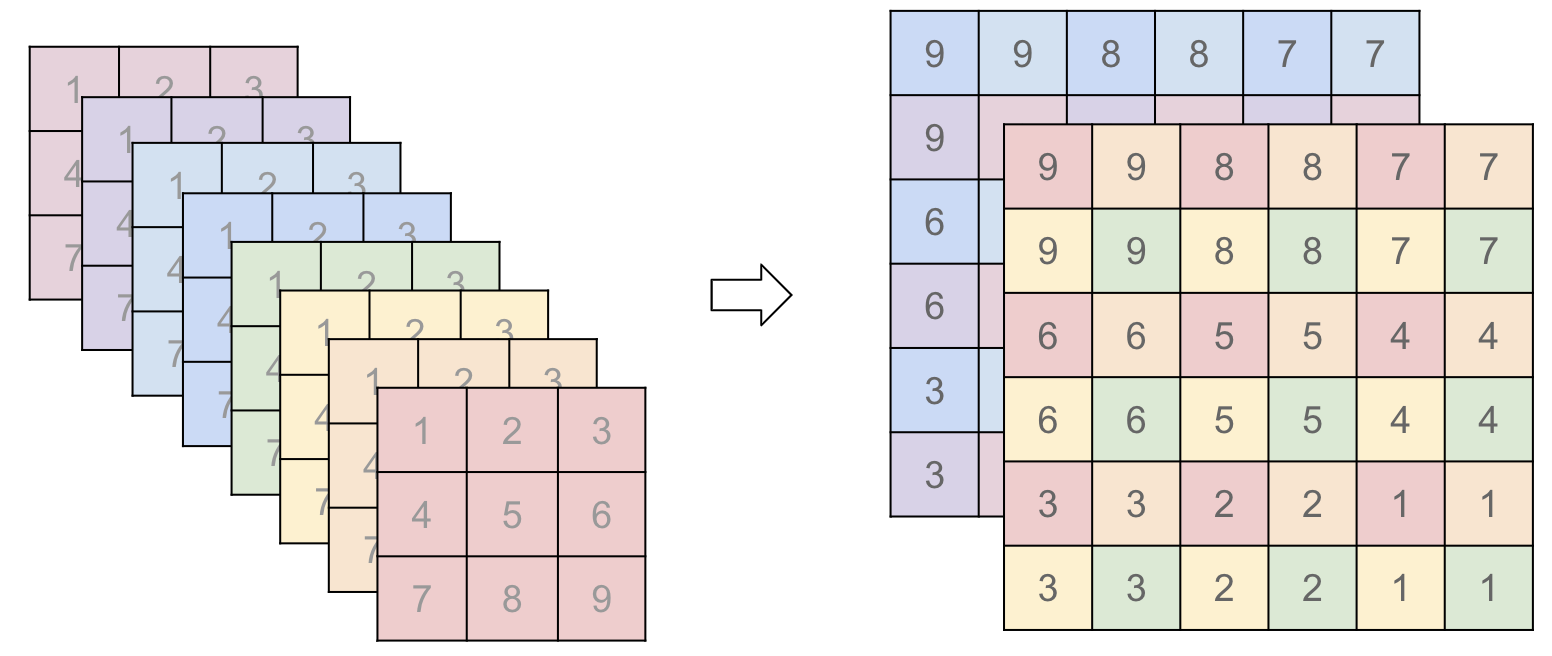}
    \caption{\small{\textbf{Weight Shuffle.} The weight shuffle algorithm moves the learned parameters of the sub-pixel convolution kernels in a similar way to the pixel shuffle algorithm, but also reverses the element indices. Unlike the pixel shuffle, this is a one-time cost in software \textit{before} deploying a trained model for inference.}}
    \label{fig:weight-shuffle}
\end{figure}

Building from the qualitative analysis of Shi \textit{et al.}~\cite{shi2016deconvolution}, we introduce the weight shuffle algorithm, given by Algorithm~\ref{alg:weight-shuffle}.
Given a sub-pixel convolution with a valid kernel size, this algorithm transforms the $K \times K$ learned convolution kernels into $rK \times rK$ deconvolution counterparts to be executed as a functionally equivalent deconvolution.
As shown in Figure~\ref{fig:weight-shuffle}, the weight shuffle algorithm moves the learned parameters of the convolution kernels in a similar way to the pixel shuffle, but also reverses element indices as a 2D matrix transpose.
Unlike the pixel shuffle algorithm, which requires a memory-intensive feature map transformation at each inference pass, the weight shuffle algorithm transforms the kernel space of a pre-trained network.
It is a one-time sunk cost that can be done in software \textit{before} deploying a trained model for inference.

% Here, the convolution kernel matrix of size $I_C \times O_C^C \times K_H^C \times K_W^C$ is transformed to a deconvolution kernel matrix of size $I_C \times O_C^D \times K_H^D \times K_W^D$ where $K^D = rK^C$ and $O_C^C = O_C^D * r^2$.
% In the context of super resolution, the kernel matrix sizes become $C \times r^2 * C \times K \times K$ and $C \times C \times r * K \times r * K$ for convolution and deconvolution, respectively.

\begin{algorithm}
\caption{Weight Shuffle}\label{alg:weight-shuffle}
\begin{algorithmic}[1]
\Procedure{Weight Shuffle}{}
\For{$i_c = 0$, $i_c{+}{+}$, while $i_c < I_C$}
\For{$o_c^d = 0$, $o_c^d{+}{+}$, while $o_c^d < O_C^D$}
\For{$k_h^d = 0$, $k_h^d{+}{+}$, while $k_k^d < K_H^D$}
\For{$k_w^d = 0$, $k_w^d{+}{+}$, while $k_w^d < K_W^D$}
\State $k_h^c \leftarrow K_H^C - \lfloor \frac{k_h^d}{r} \rfloor - 1$
\State $k_w^c \leftarrow K_W^C - \lfloor \frac{k_w^d}{r} \rfloor - 1$
\State $o_c^c \leftarrow r^2 * o_c^d~+~r~*~$\textbf{mod}$(k_h^d,r)~+~$\textbf{mod}$(k_w^d,r)$
\State kernel$_{deconv}[i_c,o_c^d,k_h^d,k_w^d$] $\leftarrow$ kernel$_{conv}[i_c,o_c^c,k_h^c,h_w^c$]
\EndFor
\EndFor
\EndFor
\EndFor
\EndProcedure
\end{algorithmic}
\end{algorithm}

% Now that we know that it \textit{can} be done, the question remains of why it \textit{should} be done.
% Next, we'll explore the algorithmic trade-offs in the context of local inference on the edge.

\subsection{Resize Convolution to Deconvolution}
\label{sec:resize-conv-to-deconv}

To upsample by a factor of $r$, the nearest neighbor (NN) resize convolution first uses NN interpolation to increase the resolution of the image before applying a same-padded convolution over the higher dimensional output space.
% As shown in Figure~\ref{fig:resize-convolution}, this results in inefficient computations as NN interpolation redundantly replicates lower-dimensional pixels only to be strided over separately by convolution kernels.
% The convolution algorithm is commonly viewed as a linear combination of the inputs $\mathbf{x}$ weighted by learned kernels $\mathbf{w}$ to give an output $y$.
% The most widely used algorithms for convolution primitives rely on this general matrix-multiple (GEMM) formulation to increase utilization of highly parallel hardware architectures such as GPUs~\cite{lavin2016fast}.
As shown below to reflect Figure~\ref{fig:resize-convolution}, the replication of each input pixel results in a unique pattern of $r^2$ identical output pixels.
Formulating the ensuing same-padded convolution as a matrix multiplication enables a significant reduction in operations.

\begin{align}
    \mathbf{y}_{0,0} ={}& \mathbf{x}_{0,0} * (\mathbf{w}_{1,1} + \mathbf{w}_{2,1} + \mathbf{w}_{1,2} + \mathbf{w}_{2,2}) = \alpha_0 \mathbf{x}_{0,0} \\
    \mathbf{y}_{0,1} ={}& \mathbf{x}_{0,0} * (\mathbf{w}_{1,0} + \mathbf{w}_{1,1} + \mathbf{w}_{2,0} + \mathbf{w}_{2,1}) + \mathbf{x}_{0,1} * (\mathbf{w}_{1,2} + \mathbf{w}_{2,2}) = \alpha_1 \mathbf{x}_{0,0} + \alpha_2 \mathbf{x}_{0,1} \\
    \mathbf{y}_{0,2} ={}& \mathbf{x}_{0,0} * (\mathbf{w}_{1,0} + \mathbf{w}_{2,0}) + \mathbf{x}_{0,1} * (\mathbf{w}_{1,1} + \mathbf{w}_{2,1} + \mathbf{w}_{1,2} + \mathbf{w}_{2,2}) = \alpha_3 \mathbf{x}_{0,0} + \alpha_0 \mathbf{x}_{0,1} \\
    \mathbf{y}_{0,3} ={}& \mathbf{x}_{0,1} * (\mathbf{w}_{1,0} + \mathbf{w}_{1,1} + \mathbf{w}_{2,0} + \mathbf{w}_{2,1}) = \alpha_1 \mathbf{x}_{0,1}
\end{align}

\noindent The pattern emerging from this reduction matches the deconvolution arithmetic in Figure~\ref{fig:deconvolution}.
\begin{align}
    \mathbf{y}_{0,0} ={}& \mathbf{x}_{0,0} \mathbf{w}^D_{1,1} \\
    \mathbf{y}_{0,1} ={}& \mathbf{x}_{0,0} \mathbf{w}^D_{1,2} + \mathbf{x}_{0,1} \mathbf{w}^D_{1,0} \\
    \mathbf{y}_{0,2} ={}& \mathbf{x}_{0,0}\mathbf{w}^D_{1,3} + \mathbf{x}_{0,1}\mathbf{w}^D_{1,1} \\
    \mathbf{y}_{0,3} ={}& \mathbf{x}_{0,1} \mathbf{w}^D_{1,2}
\end{align}

\noindent Solving for both sets of equations, we find that the linear combination follows a locally connected pattern similar to a convolution with transposed kernels, as shown below.
\begin{align}
    \mathbf{w}^D_{1,0} ={}& \mathbf{w}_{1,2} + \mathbf{w}_{2,2}\\
    \mathbf{w}^D_{1,1} ={}& \mathbf{w}_{1,1} + \mathbf{w}_{2,1} + \mathbf{w}_{1,2} + \mathbf{w}_{2,2} \\
    \mathbf{w}^D_{1,2} ={}& \mathbf{w}_{1,0} + \mathbf{w}_{1,1} + \mathbf{w}_{2,0} + \mathbf{w}_{2,1}\\
    \mathbf{w}^D_{1,3} ={}& \mathbf{w}_{1,0} + \mathbf{w}_{2,0}
\end{align}

To account for the redundant replication of input pixels, a functionally equivalent deconvolution uses a stride $S=r$ and maintains padding such that $P^D = P = \frac{K - 1}{2}$.
The resulting kernel size $K^D$ is calculated below where $K$ is the convolution kernel size given by Eq.~\ref{eq:conv_kernel_size}, $I_H$ is the input height, $S$ is the stride, and $O^H$ is the output height~\cite{dumoulin2016guide}.
$$
O_H = S \times (I_H - 1 ) + K^D - 2P^D
$$
$$
rH = r \times (H - 1) + K^D - 2\frac{K - 1}{2}
$$
$$
K^D = K + r - 1
$$

\begin{figure}[h]
    \centering
    \subfloat[2D Transpose]{\includegraphics[width=0.2\linewidth]{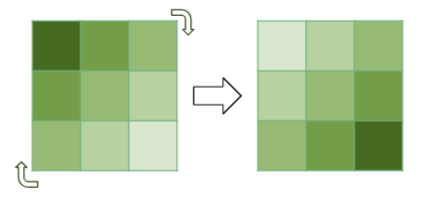}}
    ~~~~~~
    \subfloat[$t_0$]{\includegraphics[width=0.1\linewidth]{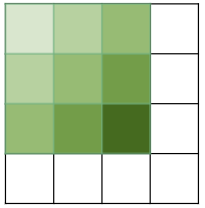}}
    ~~~~~~
    \subfloat[$t_1$]{\includegraphics[width=0.1\linewidth]{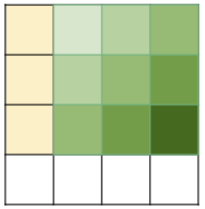}}
    ~~~~~~
    \subfloat[$t_2$]{\includegraphics[width=0.1\linewidth]{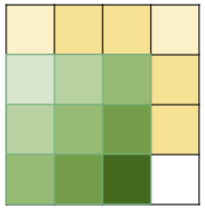}}
    ~~~~~~
    \subfloat[$t_3$]{\includegraphics[width=0.1\linewidth]{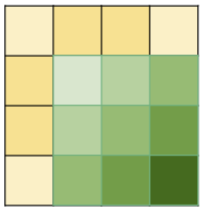}}
    ~~~~~~
    \subfloat[$t_N$]{\includegraphics[width=0.1\linewidth]{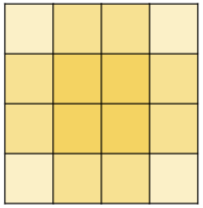}}
    \caption{\small{\textbf{Weight convolution.} In this example, we can equate the operations depicted in Figures~\ref{fig:resize-convolution} and~\ref{fig:deconvolution} by first rotating the kernels to reverse the indices, the convolving the reversed $3 \times 3$ convolution kernel (in \textcolor{ForestGreen}{green}) over the $4 \times 4$ deconvolution kernel (in \textcolor{BurntOrange}{yellow}) with a stride of 1.}}
    \label{fig:weight_convolution}
\end{figure}

Building from this algebraic reduction, we introduce the weight convolution, given by Algorithm~\ref{alg:weight-convolution}.
Given a NN resize convolutio with a valid kernel size, this algorithm transforms $K \times K$ learned convolution kernels into $(r + K - 1) \times (r + K - 1)$ deconvolution kernels to be executed as a functionally equivalent deconvolution.
To stride over the kernel space, the inequality $i + K - 1 < K^D$ must hold such that $i < r$.
As shown in Figure~\ref{fig:weight_convolution}, the weight convolution transposes the learned kernels before convolving over the weight space with a stride of 1.
Similar to the weight shuffle, this algorithm is a one-time sunk cost that is done \textit{before} deploying a trained model for inference.
However, unlike the weight shuffle, the weight convolution drastically lowers the total compute work as a consequence of the algebraic reductions.
The resulting deconvolution requires only $H\times W \times C^2 \times (r + K - 1)^2$ multiply-accumulate (MAC) operations.
When compared to the resize convolution, which requires $H\times W \times C^2 \times K^2 \times r^2$, \Note{this is a significant reduction that scales with upsampling factor $r$.}
Using the standard kernel size of 3, the resulting ratio of deconvolution MACs to NN resize convolution MACs becomes $\frac{(r + 2)^2}{(9r^2)}$.
When upsampling by a factor of 2, the resulting deconvolution only requires 44\% of the MAC operations used by its NN resize convolution counterpart to generate the same image.
When upsampling by a factor of 3, the resulting deconvolution only requires 30\% of the MACs to generate the same image.

\begin{algorithm}
\caption{Weight Convolution}\label{alg:weight-convolution}
\begin{algorithmic}[1]
\Procedure{Weight Convolution}{}
\For{$i_c = 0$, $i_c{+}{+}$, while $i_c < I_C$}
\For{$o_c = 0$, $o_c{+}{+}$, while $o_c < O_C$}
\For{$i = 0$, $i{+}{+}$, while $i < r$}
\For{$j = 0$, $j{+}{+}$, while $j < r$}
\State kernel$_{deconv}[i_c,o_c,i:i+K,j:j+K$] $\leftarrow$ kernel$_{conv}[i_c,o_c,:,:$]$^T$
\EndFor
\EndFor
\EndFor
\EndFor
\EndProcedure
\end{algorithmic}
\end{algorithm}

% \begin{figure}[h]
%     \centering
%     \includegraphics[width=0.4\linewidth]{figs/mac-reduction.png}
%     \caption{\textbf{Reducing MAC operations with the weight convolution.} Using the weight convolution algorithm to translate a NN resize convolution to a deconvolution reduces the overall MACs significantly as the scaling factor $r$ increases.}
%     \label{fig:mac-reduction}
% \end{figure}

\section{Time and Energy Analysis for Local Inference}
\label{sec:experimental-results}

In our proposed edge computing paradigm depicted in Figure~\ref{fig:main_idea}, a deep learning model is first trained in the cloud using either sub-pixel convolution (C-SP) or nearest neighbor resize convolution (C-NN).
The learned weights are then recast by kernel transformations (Section~\ref{sec:translation-algos}) to be deployed for inference as a functionally equivalent deconvolution - D-SP or D-NN, respectively.
In Section~\ref{sec:deconvolution}, we discuss the various formulations of deconvolution that avoid the overlapping summation problem observed when the kernel size $K$ is greater than the stride $S$.
We consider the following deconvolution variants as \Note{low-level} inference algorithms - improved reverse looping deconvolution (REVD2), transforming deconvolution to convolution (TDC), and fractionally strided deconvolution (STRD).
% We present a quantitative analysis to verif the selection of REVD2 among these backend deconvolution algorithms in the context of time and energy.
Here, we present a framework for quantitative analyses to validate our proposed paradigm based on the properties of each algorithm.
Results and conclusions are discussed in Section~\ref{sec:discussion}.

\subsection{Quantitative Models for Time and Energy Costs}
\label{sec:time_and_energy_costs}

We use the simplified \textit{optimistic} quantitative model detailed by~\cite{choi2013roofline} to compare the time and energy costs of each algorithm characterized by their compute and memory requirements\footnote{The assumptions held by this model yield a best-cast analysis which is only valid when algorithms are sufficiently parallelizable~\cite{choi2013roofline}. As discussed in Section~\ref{sec:deconvolution}, each of the algorithms considered are data-independent along the output space and can therefore be executed with high degrees of concurrency.}.
Given that $T_\text{comp}$ and $T_\text{mem}$ are the total time to execute all compute and memory operations, respectively, the time cost model given by Equation~\ref{eq:time_cost} assumes an idealized hardware design that perfectly masks the communication overhead with computation work~\cite{choi2013roofline}.
Here, a higher computation-to-communication ratio would better hide memory bandwidth bottlenecks.
\begin{equation}
T \equiv \text{max}(T_\text{comp}, T_\text{mem})
\label{eq:time_cost}
\end{equation}

\noindent Similarly, let $E_\text{comp}$ and $E_\text{mem}$ be the total energy to execute all compute and memory operations, respectively, and let $E_0(T)$ be the cost of constant energy expended while executing the algorithm.
Unlike time cost, the energy cost model given by Equation~\ref{eq:energy_cost} does not overlap computation and communication costs and has an additional penalty for increased latency~\cite{choi2013roofline}.
\begin{equation}
E \equiv E_\text{comp} + E_\text{mem} + E_0(T)
\label{eq:energy_cost}
\end{equation}

\noindent For a fixed hardware architecture, let $\tau_\text{comp}$ and $\tau_\text{mem}$ be the time cost per compute and memory operation, respectively.
For a given algorithm, let $C$ be the total number of computation operations and $M$ the total number of memory operations required.
Under the optimistic assumption of hiding memory latency with perfect overlap, the total time cost of an algorithm becomes Equation~\ref{eq:time_cost_algo}, where $T_\text{comp} \equiv C \tau_\text{comp}$ and $T_\text{mem} \equiv M \tau_\text{mem}$~\cite{choi2013roofline}.
\begin{equation}
T = \text{max}(C \tau_\text{comp}, M \tau_\text{mem})
\label{eq:time_cost_algo}
\end{equation}

\noindent Similarly, let $\epsilon_\text{comp}$ and $\epsilon_\text{mem}$ be the energy cost per compute and memory operation, respectively.
The total energy cost of an algorithm then becomes Equation~\ref{eq:energy_cost_algo}, where $E_\text{comp} \equiv C \epsilon_\text{comp}$, $E_\text{mem} \equiv M \epsilon_\text{mem}$, and the constant energy cost is assumed to be linear in time with a fixed constant power defined by $\pi_0$ such that $E_0(T) \equiv \pi_0 T$~\cite{choi2013roofline}.
\begin{equation}
E \equiv C \epsilon_\text{comp} + M \epsilon_\text{mem} + \pi_0 T
\label{eq:energy_cost_algo}
\end{equation}

\noindent Given the compute and memory requirements of an algorithm, we use Equations~\ref{eq:time_cost_algo} and~\ref{eq:energy_cost_algo} to estimate time and energy costs using this idealized abstraction of hardware performance.

\subsubsection{Deconvolution Time and Energy Costs}

\begin{figure}[t]
    \centering
    \subfloat[D-SP: Time Cost]{\includegraphics[width=0.47\linewidth]{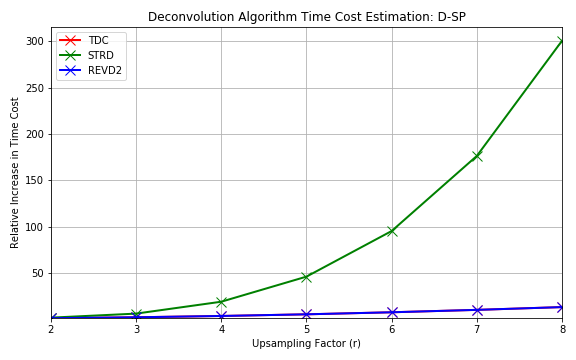}}
    ~
    \subfloat[D-NN: Time Cost]{\includegraphics[width=0.47\linewidth]{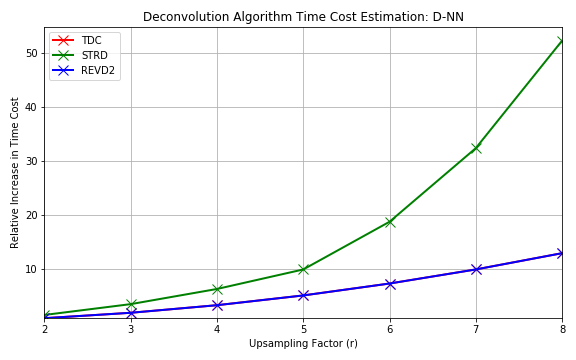}} \\
    \subfloat[D-SP: Energy Cost]{\includegraphics[width=0.47\linewidth]{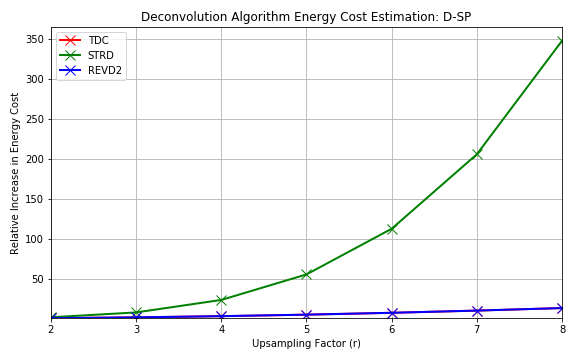}}
    ~
    \subfloat[D-NN: Energy Cost]{\includegraphics[width=0.48\linewidth]{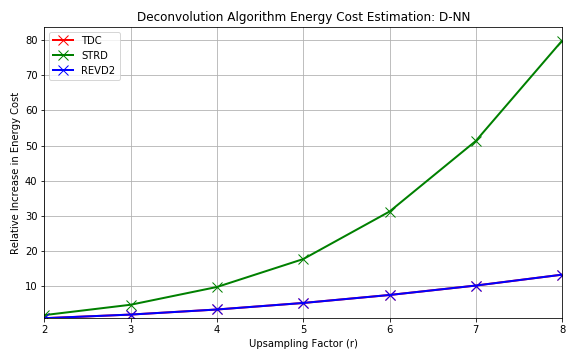}} \\
    \caption{\small{\textbf{Deconvolution Time and Energy Cost Estimation.} Using the \textit{optimistic} quantitative model proposed in~\cite{choi2013roofline}, we analyze the relative increase in time and energy costs increase as a function of upscaling factor $r$.
    Each experiment assumes a square 1K RGB input image upsampled using a standard $3 \times 3$ kernel on NVIDIA's GeForce GTX 680.
    We normalize all values by REVD2 time and energy costs without upsampling ($r=1$). As shown in Table~\ref{tbl:mem-requirements-algos}, the compute ($C$) and activation ($A$) requirements of TDC and REVD2 are equal. With data movement dominated by activations ($A$), any variance in weights ($W$) is minimally impactful.}}
    \label{fig:time_energy_cost_deconv_algos}
\end{figure}

Unlike REVD2, both STRD and TDC require the insertion of zeros to upsample an image by a factor of $r$.
As shown in Figure~\ref{fig:sparsity_requirements}, the presence of these redundant zero-valued computations increases with the upsampling factor and cannot be ignored when analyzing data movement patterns\footnote{Zero-skipping techniques can lessen the impact of increased sparsity. However, it requires control logic that can introduce overhead and, if not properly balanced across concurrent processes, it can also introduce synchronization issues in multi-threaded hardware~\cite{biookaghazadeh2018fpgas,nurvitadhi2017can}. We ignore the impact of these optimizations in our quantitative analysis.}.
% While techniques such as zero-skipping lessen the impact of increased sparsity,
Table~\ref{tbl:mem-requirements-algos} summarizes the compute and memory requirements for each of the \Note{low-level} deconvolution algorithms.
Compute requirements ($C$) are measured by the number of multiply-accumulates (MACs).
Memory requirements ($M$) are measured by the number of weights ($W$) and activations ($A$), \textit{i.e.} the sum of the input and output feature maps such that $M \equiv W + A$.
We separately consider deconvolution as translated from sub-pixel convolution (D-SP) and deconvolution as translated from nearest neighbor resize convolution (D-NN).
To estimate time and energy costs, we translate these compute and memory requirements using the idealized abstraction of hardware performance given by Equations~\ref{eq:time_cost_algo} and~\ref{eq:energy_cost_algo}.
In each experiment, we consider the case of upsampling a square 1K RGB image by a factor of $r$ using $3 \times 3$ kernels\footnote{To simplify analyses, we alter the standard 1K RGB image resolution, which is $1024 \times 768 \times 3$,  to be square such that $1024 \times 1024 \times 3$.}.
We assume all pixel values and network parameters are executed and stored at 32-bit precision, \textit{i.e.} 4 bytes.
Figure~\ref{fig:time_energy_cost_deconv_algos} shows the relative increase in time and energy costs as a function of upsampling factor $r$ for each \Note{low-level} deconvolution algorithm using either D-SP or D-NN formulations when executing on the NVIDIA GeForce GTX 680\footnote{Choi~\textit{et al.}~\cite{choi2013roofline} provide the values for $\tau_\text{comp}, \tau_\text{mem}, \epsilon_\text{comp}, \epsilon_\text{mem}$ for NVIDIA's GeForce GTX 680, NIVIDA's GeForce GTX 580, and Intel's Core i7-950.}.
Because memory requirements are dominated by activations rather than weights, as shown in Table~\ref{tbl:mem-requirements-algos}, the impact of the zero-insertion requirements for STRD are massive while those for TDC are minimal.
As discussed in Section~\ref{sec:revd2}, there are no zero-insertion requirements for REVD2.

\begin{table}[h]
\centering
\begin{tabular}{lccc}
\hline
\multicolumn{1}{c}{D-SP} & \textbf{\#MACs ($C$)} & \textbf{\# Parameters ($W$)} & \textbf{\# Activations ($A$)} \\ \hline
\multicolumn{1}{|l|}{\textbf{REVD2}} & \multicolumn{1}{c|}{$~~~~r^2 \times K^2 \times H^2 \times C^2~~~~~~~~$} & \multicolumn{1}{c|}{$r^2 \times K^2 \times C^2$} & \multicolumn{1}{c|}{$(1 + r^2) \times H^2 \times C$} \\ \hline
\multicolumn{1}{|l|}{\textbf{STRD}} & \multicolumn{1}{c|}{$~~~~r^4 \times K^2 \times H^2 \times C^2~~~~~~~~$} & \multicolumn{1}{c|}{$r^2 \times K^2 \times C^2$} & \multicolumn{1}{c|}{$\left(r^2 \times H^2 + (H + P_H)^2\right) \times C$} \\ \hline
\multicolumn{1}{|l|}{\textbf{TDC}} & \multicolumn{1}{c|}{$~~~~r^2 \times K^2 \times H^2 \times C^2~~~~~~~~$} & \multicolumn{1}{c|}{$r^2 \times K^2 \times C^2$} & \multicolumn{1}{c|}{$(1 + r^2) \times H^2 \times C$} \\ \hline
\end{tabular}
\begin{tabular}{lccc}
\multicolumn{1}{c}{D-NN} & \textbf{\#MACs ($C$)} & \textbf{\# Parameters ($W$)} & \textbf{\# Activations ($A$)} \\ \hline
\multicolumn{1}{|l|}{\textbf{REVD2}} & \multicolumn{1}{c|}{$r^2 \times \lceil\frac{r + K - 1}{r}\rceil^2 \times H^2 \times C^2 $} & \multicolumn{1}{c|}{$(r + K - 1)^2 \times C^2$} & \multicolumn{1}{c|}{$(1 + r^2) \times H^2 \times C$} \\ \hline
\multicolumn{1}{|l|}{\textbf{STRD}} & \multicolumn{1}{c|}{$r^2 \times (r + K - 1)^2 \times H^2 \times C^2 $} & \multicolumn{1}{c|}{$(r + K - 1)^2 \times C^2$} & \multicolumn{1}{c|}{$\left(r^2 \times H^2 + (H + P_H)^2\right) \times C$} \\ \hline
\multicolumn{1}{|l|}{\textbf{TDC}} & \multicolumn{1}{c|}{$r^2 \times \lceil\frac{r + K - 1}{r}\rceil^2 \times H^2 \times C^2 $} & \multicolumn{1}{c|}{$r^2 \times \lceil\frac{r + K - 1}{r}\rceil^2 \times C^2$} & \multicolumn{1}{c|}{$(1 + r^2) \times H^2 \times C$} \\ \hline
\end{tabular}
\caption{\small{\textbf{Deconvolution Compute and Memory Requirements.} For simplicity, we assume square kernels $K$, square inputs $H$, and equal input/output channels $C$.
We define $P_H = (H-1)(r-1)$ as the zeros inserted to pad each input pixel for the fractionally strided deconvolution (STRD).}}
% Note that, for image upsampling, $r \geq 2$. For D-NN, it follows that $\lceil \frac{K}{S} \rceil = \lceil \frac{r + 2}{r} \rceil = \lceil 1 + \frac{2}{r} \rceil = 2$}}
\label{tbl:mem-requirements-algos}
\end{table}

\subsubsection{Convolution-based Upsampling Algorithm Time and Energy Costs}

\begin{figure}[h]
    \centering
    \subfloat[Time Cost]{\includegraphics[width=0.5\linewidth]{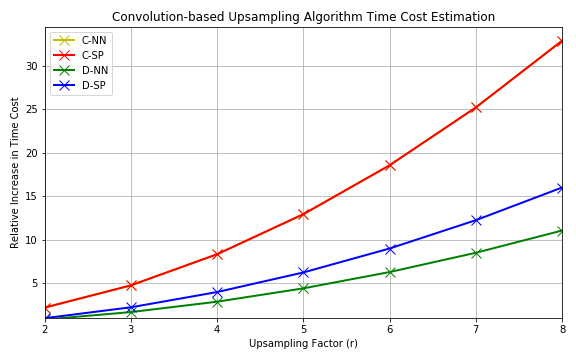}}
    ~
    \subfloat[Energy Cost]{\includegraphics[width=0.5\linewidth]{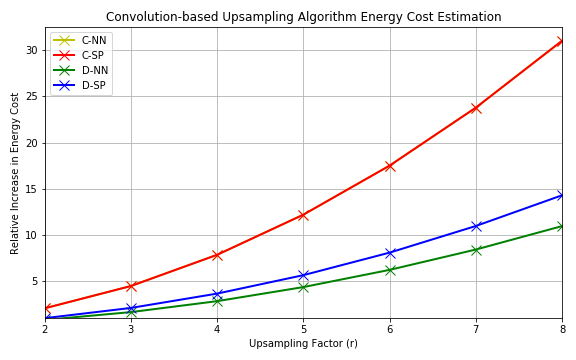}}
    \caption{\small{\textbf{Convolution-based Upsampling Algorithm Time and Energy Cost Estimation.} Using the \textit{optimistic} quantitative model proposed in~\cite{choi2013roofline}, we analyze the relative increase in time and energy costs as a function of upsampling factor $r$. Each experiment assumes a square 1K RGB input image upsampled using a standard $3 \times 3$ kernel on NVIDIA's GeForce GTX 680. We normalize all values by D-SP time and energy costs without upsampling ($r=1$). As shown in Table~\ref{tbl:mem-requirements}, the compute ($C$) and activation ($A$) requirements of C-SP and C-NN are equal. With data movement dominated by activations ($A$), any variance in weights ($W$) is minimally impactful. We assume REVD2 as the \Note{low-level} deconvolution algorithm.}}
    \label{fig:conv_ops_time_energy}
\end{figure}

% We apply these insights to analyze the energy efficiency of each of the convolution-based upsampling algorithms.
Whereas deconvolution directly upsamples an image in one operation, both the sub-pixel convolution (C-SP) and nearest neighbor resize convolution (C-NN) rely on memory-dominated operations to move data for post- and pre-processing, respectively.
These operations are required for every inference pass and cannot be ignored when analyzing data movement patterns.
Table~\ref{tbl:mem-requirements} summarizes the compute and memory requirements for each of the \Note{high-level} convolution-based image upsampling algorithms.
Note that the compute ($C$) and activation ($A$) requirements of C-SP and C-NN are equal.
The sub-pixel convolution performs computations in low resolution (LR) space but generates $r^2$ more output channels and requires expensive post-processing in high resolution (HR) space.
The resize convolution requires less-expensive pre-processing in LR space, but performs its computations in HR space.
Using kernel transformations to enable deconvolution for inference at the edge avoids any penalties for expensive data pre- or post-processing while maintaining the image fidelity learned through training in the cloud.
We separately consider deconvolution as translated from sub-pixel convolution (D-SP) and deconvolution as translated from nearest neighbor resize convolution (D-NN).
For each experiment, we assume REVD2 as the \Note{low-level} algorithm and consider the case of upsampling a square 1K RGB image by a factor of $r$ using standard kernels of size $3 \times 3$.
We assume all pixel values and network parameters are executed and stored at 32-bit precision.
Figure~\ref{fig:conv_ops_time_energy} shows the relative increase in time and energy costs as a function of upscaling factor $r$ when executing each convolution-based upsampling algorithm on the NVIDIA GeForce GTX 680.
With activations dominating memory requirements, any deviation in weight requirements ($W$) has minimal impact on time and energy costs.

\begin{table}[t]
\centering
\begin{tabular}{lccc}
\hline
\multicolumn{1}{c}{} & \textbf{\#MACs ($C$)} & \textbf{\# Parameters ($W$)} & \textbf{\# Activations ($A$)} \\ \hline
\multicolumn{1}{|l|}{\textbf{C-SP}}   & \multicolumn{1}{c|}{$r^2 \times K^2 \times H^2 \times C^2$} & \multicolumn{1}{c|}{$r^2 \times K^2 \times C^2$} & \multicolumn{1}{c|}{$(1 + 3r^2) \times H^2 \times C$} \\ \hline
\multicolumn{1}{|l|}{\textbf{C-NN}} & \multicolumn{1}{c|}{$r^2 \times K^2 \times H^2 \times C^2$} & \multicolumn{1}{c|}{$K^2 \times C^2 $} & \multicolumn{1}{c|}{${(1 +  3r^2) \times H^2 \times C}$} \\ \hline
\multicolumn{1}{|l|}{\textbf{D-SP}}        & \multicolumn{1}{c|}{$r^2 \times K^2 \times H^2 \times C^2$} & \multicolumn{1}{c|}{$r^2 \times K^2 \times C^2$} & \multicolumn{1}{c|}{$(1 + r^2) \times H^2 \times C$} \\ \hline
\multicolumn{1}{|l|}{\textbf{D-NN}}       & \multicolumn{1}{c|}{$r^2 \times \lceil\frac{r + K - 1}{r}\rceil^2 \times H^2 \times C^2 $} & \multicolumn{1}{c|}{$(r + K - 1)^2 \times C^2$} & \multicolumn{1}{c|}{$(1 + r^2) \times H^2 \times C$} \\ \hline
\end{tabular}
\caption{\small{\textbf{Convolution-based Upsampling Algorithm Compute and Memory Requirements.} For simplicity, we assume square kernels $K$, square inputs $H$, and equal input/output channels $C$. Note that both C-SP and C-NN activations include the pixel shuffle and NN-interpolation, respectively, as they are required for every inference pass. We assume REVD2 as the \Note{low-level} deconvolution algorithm.}} % for D-SP and D-NN.}}
\label{tbl:mem-requirements}
\end{table}

\subsection{Estimating Energy Efficiency using Data Reuse Patterns}
\label{sec:time_and_energy_efficiency}

The efficiency of an algorithm is typically described by data reuse and measured using arithmetic intensity - the number of \textit{useful} compute operations for every byte of data accessed.
Algorithms with high data reuse are more likely to yield performance improvements with an increase in compute resources because computations dominate the communication overhead.
% a high ratio implies operations are dominated by compute work so each byte of data accessed is more frequently reused.
Algorithms with low data reuse put more strain on a system's memory bandwidth as each compute operation requires more off-chip memory accesses.
% The intensity of an algorithm is inherently independent of hardware memory capacity as it measures the locality of an algorithm~\cite{choi2013roofline}.
% Data reuse is typically measured using arithmetic intensity (AI), defined as the ratio of the total compute operations to the total bytes accessed.
% However, energy consumption is dominated by data movement rather than computation~\cite{horowitz20141}.
While the value of this ratio implies the scalability and locality of an algorithm, it fails to properly estimate the energy efficiency of convolution-based deep learning algorithms when the ratio of activations ($A$) to weights ($W$) drastically deviates from 1~\cite{jha2020hardware}.
By separately considering weight and activation reuse, Jha \textit{et al.}~\cite{jha2020hardware} show that, for convolution-based deep learning algorithms, the variation in arithmetic intensity is attributed to the variation in activation reuse and is highly correlated to variations in energy efficiency\footnote{Energy efficiency is typically measured in units of GFLOPs per Joule as the analog to time efficiency, \textit{e.g.} throughput.}.
% they propose a data type aware weighted arithmetic intensity ($DI_\alpha$) given by Equation~\ref{eq:modified_arithmetic_intensity}, where $C$ is the number of \textit{useful} compute operations, $A$ is the bytes of activations accessed, $W$ is bytes of weights accessed, and $\alpha$ is a data reuse coefficient that can be tuned to maximize the correlation with energy efficiency.
% \begin{equation}
%     DI_\alpha = \frac{1}{4} \left[ \alpha * \frac{C}{A} + (1 - \alpha) * \frac{C}{W} \right]
%     \label{eq:modified_arithmetic_intensity}
% \end{equation}
% For simplicity of analysis and visualization, we this modified arithmetic intensity ($DI_\alpha$) as our proxy for data reuse and set $\alpha=1$ to focus solely on the activation reuse.
Following this work, we use the compute and memory requirements discussed in Section~\ref{sec:time_and_energy_costs} to estimate the energy efficiencies of convolution-based upsampling algorithms using activation reuse.
% To compare convolution-based upsampling operations for energy efficiency inference, we analyze the intensity of the sub-pixel convolution (C-SP) and NN resize convolution (C-NN) against their deconvolution counterparts (D-SP \& D-NN, respectively).

% \subsubsection{Data Reuse Patterns for Back-end Deconvolution Algorithms}

\begin{figure}[h]
    \centering
    \subfloat[]{\includegraphics[width=0.33\linewidth]{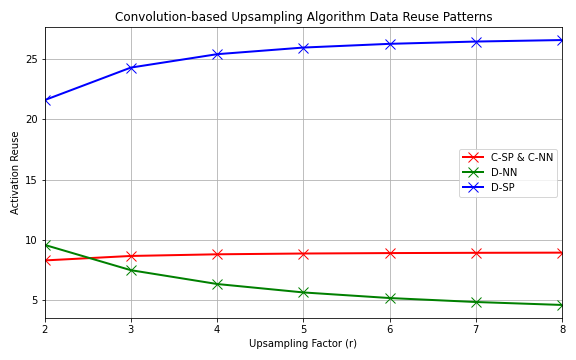}}
    \subfloat[D-SP]{\includegraphics[width=0.33\linewidth]{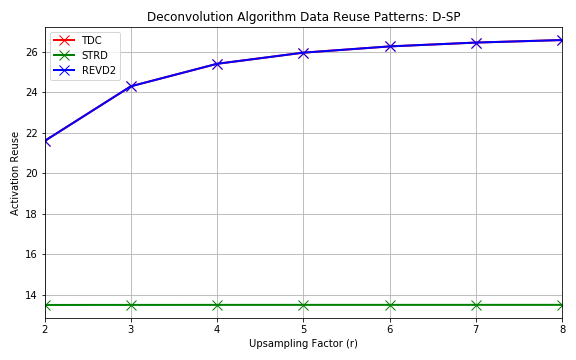}}
    \subfloat[D-NN]{\includegraphics[width=0.33\linewidth]{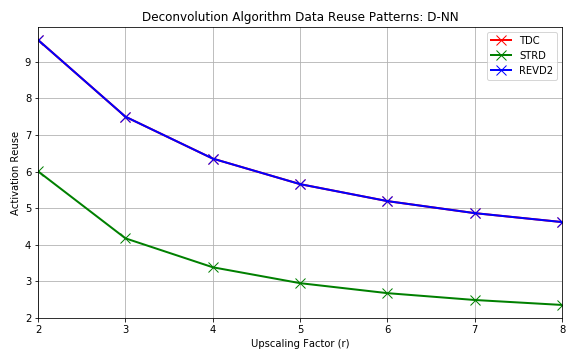}}
    \caption{\small{\textbf{Effect of Upsampling Factor on Activation Reuse.} We use the compute and memory requirements from Tables~\ref{tbl:mem-requirements-algos} and~\ref{tbl:mem-requirements} to calculate the activation reuse for each convolution-based image upsampling algorithm. Following the work of Jha \textit{et al.}~\cite{jha2020hardware}, we use this metric to estimate the energy efficiency of each algorithm as a function of upsampling factor $r$. Each experiment assumes a square 1K RGB input image upsampled using a standard $3\times3$ kernel.}}
    \label{fig:activation_reuse}
\end{figure}

When estimating the energy efficiency using activation reuse, we define \textit{useful} compute operations as those contributing to output pixel values, \textit{i.e.} non-zero-valued computations.
However, energy consumption is dominated by data movement rather than computation~\cite{horowitz20141}.
As such, we define activation reuse as the number of non-zero-valued computations for every byte of activation data accessed when upsampling an image by a factor of $r$.
Figure~\ref{fig:activation_reuse} shows how activation reuse increases with upsampling factor $r$ for each convolution-based upsampling algorithm.
For D-SP and D-NN, we assume REVD2 as the \Note{low-level} deconvolution algorithm.
Again, we consider the case of upsampling a square 1K RGB image by a factor of $r$ using standard kernels of size $3 \times 3$ where all pixel values and network parameters are executed and stored at 32-bit precision.

\subsection{Roofline Models of Time and Energy}
\label{sec:roofline_models}

\begin{figure}
    \centering
    \subfloat[]{\includegraphics[width=0.33\linewidth]{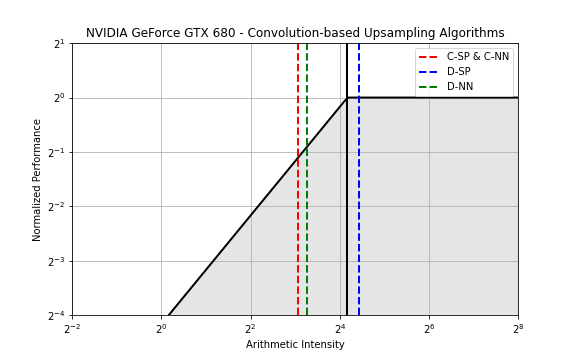}}
    \subfloat[]{\includegraphics[width=0.33\linewidth]{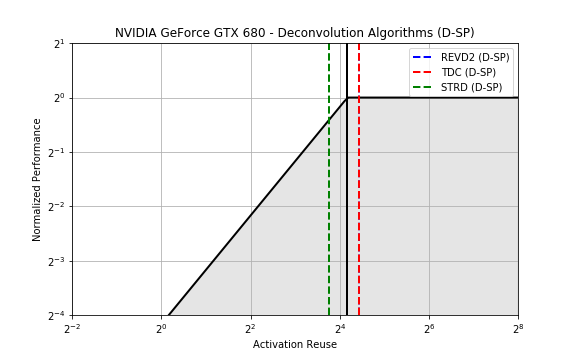}}
    \subfloat[]{\includegraphics[width=0.33\linewidth]{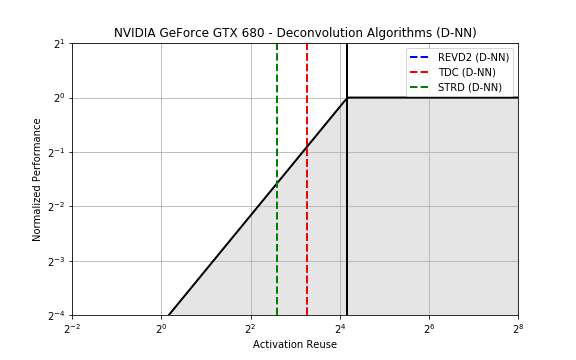}} \\
    \subfloat[]{\includegraphics[width=0.33\linewidth]{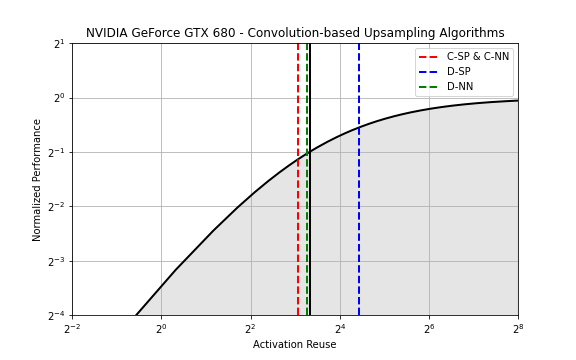}}
    \subfloat[]{\includegraphics[width=0.33\linewidth]{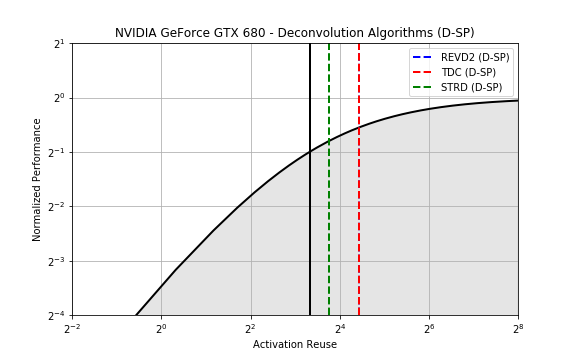}}
    \subfloat[]{\includegraphics[width=0.33\linewidth]{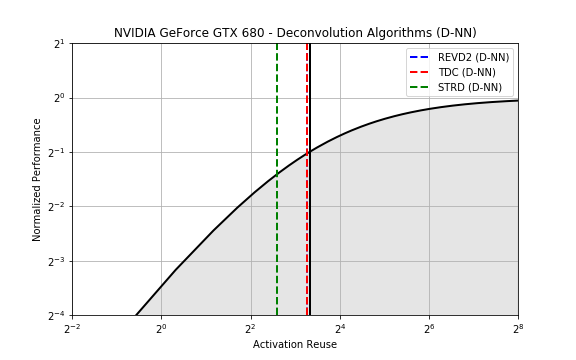}}
    % \subfloat[]{\includegraphics[width=0.35\linewidth]{figs/conv-upsample-ops-roofline-i7-950-D-NN.png}}
    % \subfloat[]{\includegraphics[width=0.35\linewidth]{figs/conv-upsample-ops-roofline-GTX-580-D-NN.png}}

    % \subfloat[]{\includegraphics[width=0.5\linewidth]{figs/energy-balance-scaling-factor.png}}
    \caption{\small{\textbf{Normalized Roofline Models of Time and Energy.} Here, we visualize roofline models of time (top row) and energy (bottom row) for NVIDIA's GeForce GTX 680. For each experiment, the \textbf{black} curve is the roofline model and the vertical \textbf{black} line is its respective balance point using values provided by~\cite{choi2013roofline}. The roofs of each model are normalized to peak performance. As discussed in Section~\ref{sec:time_and_energy_efficiency}, we use arithmetic intensity for the time roofline model~\cite{williams2009roofline} and activation reuse for the energy roofline model~\cite{choi2013roofline,jha2020hardware}.}}
    \label{fig:roofline}
\end{figure}

The hardware architecture analog to arithmetic intensity is time-balance~\cite{choi2013roofline}.
For a fixed machine, its time-balance point ($B_\tau$) is defined as the ratio of its time cost per memory operation ($\tau_\text{mem}$) to its time cost per compute operation ($\tau_\text{comp}$) such that $B_\tau \equiv \tau_\text{mem} / \tau_\text{comp}$~\cite{choi2013roofline}.
Similarly, the energy-balance point ($B_\epsilon$) of a machine is defined as the ratio of its energy cost per memory operation ($\epsilon_\text{mem}$) to its energy cost per compute operation ($\epsilon_\text{comp}$) such that $B_\epsilon \equiv \epsilon_\text{mem} / \epsilon_\text{comp}$~\cite{choi2013roofline}.
When the arithmetic intensity of an algorithm is equal to the balance of a machine such that $\text{AI} = B$, the cost of the algorithm's compute operations is equal to that of its memory operations.
% \Note{The design goal is to create algorithms with high intensities relative to these balance points.}
We can visualize these balance principles using roofline models of time~\cite{williams2009roofline} and energy~\cite{choi2013roofline}, which provide an upper bound on the attainable performance of an algorithm for a fixed hardware as a function of that algorithm's data reuse patterns.
Figure~\ref{fig:roofline} shows these roofline models using the balance points provided by~\cite{choi2013roofline}.
Here, we consider the case of upsampling a square 1K RGB image by a factor of 2 using $3 \times 3$ kernels at 32-bit precision.
The roofs of each model are normalized to peak performance and data reuse is measured by arithmetic intensity for time and activation reuse for energy.

\section{Results of Quantitative Analysis}
\label{sec:discussion}

% Cloud computing systems can have nearly limitless resources, making them ideal for resource-intensive tasks such as data storage, data processing, and model training.
% However, for latency-sensitive deep learning applications, edge computing paradigms decrease system latency and improve overall security by executing inference locally on edge devices without reliance on a stable network connection~\cite{dhar2019device, sze2017efficient}.
Local edge devices are often limited by battery life and hardware area which constrains the power budget and on-chip resources available for inference~\cite{dhar2019device, sze2017efficient}.
To support real-time deep learning applications, low latency and high energy efficiency become critical.
Using the quantitative models detailed in Section~\ref{sec:experimental-results}, we analyze and compare the properties of convolution-based upsampling algorithms using metrics of time and energy.
Real-time image upsampling applications, such as single-image super resolution, typically process images sequentially in batch sizes of 1.
When considering such applications, we define latency as the time cost incurred by upsampling a single image.
We evaluate energy efficiency under two metrics.
First, we use energy per pixel to evaluate the energy cost incurred by upsampling a single image~\cite{jha2020hardware}.
This is defined as the total energy cost divided by the total output pixels generated and is measured in units of Joules/pixel~\cite{jha2020hardware}.
Second, we use performance per energy to evaluate the rate of computation for every unit of energy consumed.
This is defined as the total useful computations divided by the total energy cost and, as the energy analog of throughput, is measured in units of MACs/Joule~\cite{choi2013roofline}.
Using these metrics, we validate the use of kernel transformations in our proposed edge computing paradigm assuming REVD2 as the \Note{low-level} deconvolution algorithm.
	
As shown in Table~\ref{tbl:mem-requirements}, the compute ($C$) and activation requirements ($A$) of the sub-pixel convolution (C-SP) and nearest neighbor resize convolution (C-NN) are equal\footnote{For simplicity of analysis, we ignore the impact of address calculations and modulo arithmetic.}.
With activations dominating memory requirements ($M$) and, therefore, data transfer penalties, the variations in weight requirements ($W$) have negligible impact on time and energy costs.
By translating their learned kernels for deconvolution inference, the kernel transformations introduced in Section~\ref{sec:translation-algos} remove the reliance of C-SP and C-NN on the memory-intensive feature map transformations that increase activate requirements.
We highlight the following implications of this reduction in memory accesses, assuming REVD2 as the \Note{low-level} deconvolution inference algorithm. \\

\begin{figure}[h]
    \centering
    \subfloat[]{\includegraphics[width=0.5\linewidth]{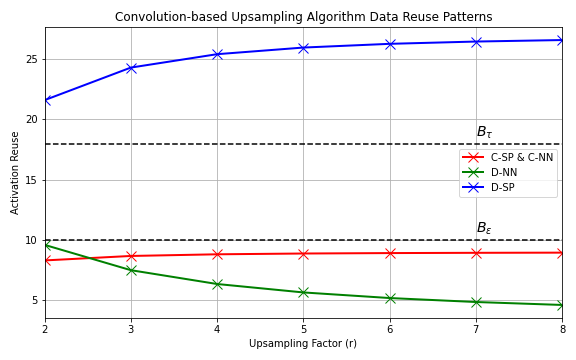}}
    \caption{\small{\textbf{The Scalability of Convolution-based Upsampling Algorithms.} The memory-intensive feature map transformations of C-SP and C-NN render each algorithm memory-bound in both time and energy on NVIDIA's GeForce GTX 680. Avoiding the pixel shuffle post-processing of C-SP drastically increases the efficiency of D-SP for inference. The increased activation reuse implies increased time and energy efficiency as compute resources scale because the overwhelming majority of work is dedicated to computations rather than memory accesses~\cite{choi2013roofline}. For D-NN, the significant reduction in MACs outweighs the reduced memory accesses, ultimately rendering it memory-bound in both time and energy as the memory pressure increases faster than the compute workload as upsampling factor increases. Unlike C-SP, C-NN, and even D-NN, D-SP remains compute-bound in both time and energy as upsampling factor increases.}}
    \label{fig:scalability}
\end{figure}

\begin{enumerate}
    \item \textit{Alleviating pressure on memory bandwidth significantly reduces system latency.}
    The detrimental impact of memory-intensive feature map transformations on C-SP and C-NN increases with upsampling factor.
    Translating C-SP to D-SP removes its reliance on the pixel shuffle post-processing and, as discussed in Section~\ref{sec:translation-algos}, translating C-NN to D-NN significantly reduces the total MACs required when removing its reliance on interpolation pre-processing.
    Figure~\ref{fig:conv_ops_time_energy}a shows that the time cost of each algorithm increases with upsampling factor $r$ using the optimistic computational model detailed in Section~\ref{sec:time_and_energy_costs}.
    \Note{In our experiments,} upsampling an image by a factor of 2 using D-SP decreases system latency by 2.2x when compared to C-SP and using D-NN decreases system latency by 2.6x when compared to C-NN.

    \item \textit{Reducing activation requirements and, therefore, data transfers significantly reduces the energy cost of upsampling an image.}
    Energy consumption is dominated by data movement~\cite{horowitz20141}.
    With activation requirements dominating memory accesses, removing the memory-intensive feature map transformations of C-SP and C-NN reduces the energy consumed when generating an output image.
    Figure~\ref{fig:conv_ops_time_energy}b shows how the energy cost of each algorithm increases with upsampling factor $r$ using the optimistic computational model detailed in Section~\ref{sec:time_and_energy_costs}.
    As each algorithm is generating an $rH \times rH \times C$ output image, we interpret this as the relative increase in energy per pixel.
    \Note{In our experiments,} upsampling an image by a factor of 2 using D-SP decreases the energy per pixel by 2.1x when compared to C-SP and using D-NN decreases the energy per pixel by 2.5x when compared to C-NN.

    \item \textit{Reducing data transfers significantly improves the rate of computation for every unit of energy consumed.}
    As discussed in Section~\ref{sec:time_and_energy_efficiency}, activation reuse is defined as the ratio of useful compute work to activation requirements.
    The activation reuse of convolution-based upsampling algorithms is tightly correlated with performance per energy (PPE)~\cite{jha2020hardware}.
    Figure~\ref{fig:activation_reuse} shows how the activation reuse of each convolution-based upsampling algorithm increases with upsampling factor $r$.
    While removing reliance on memory-intensive feature map transformations improves the PPE of D-SP, it reduces the PPE of D-NN as the reduction in MACs significantly outweighs the reduction in memory accesses.
    This imbalanced reduction ultimately renders D-NN memory-bound as the amount of compute work grows slower than the amount of memory accessed.

    \item \textit{Reducing memory accesses improves algorithm scalability.}
    The ratio of useful computations to memory accesses, \textit{i.e.} arithmetic intensity, implies the scalability of an algorithm.
    As described in Section~\ref{sec:roofline_models}, algorithms with an arithmetic intensity lower than the machine balance point are ultimately bound by memory bandwidth~\cite{williams2009roofline}.
    Algorithms with an arithmetic intensity higher than the machine balance point are ultimately bound by compute resources and are more likely to see performance gains as resources scale~\cite{choi2013roofline}.
    For each convolution-based upsampling algorithm, we use the roofline models depicted in Figure~\ref{fig:roofline} to visualize the relationships of their arithmetic intensities to the time and energy machine balance points of NVIDIA's GeForce GTX 680.
    We further show how these relationships change with upsampling factor in Figure~\ref{fig:scalability}.
    Unlike C-SP, C-NN, and even D-NN, D-SP remains compute-bound in both time and energy as upsampling factor increases.
    With compute work dominating memory accesses, the increased arithmetic intensity of D-SP implies increased time and energy efficiency as compute resources scale~\cite{choi2013roofline}.
\end{enumerate}

These trends do not hold for all selections of deconvolution formulations.
Deep learning frameworks commonly use the fractionally strided deconvolutionn (STRD) formulation to leverage unmodified convolution accelerators~\cite{NEURIPS2019_9015, tensorflow2015-whitepaper}.
However, the zero-insertion requirements on the input feature maps exponentially increase the data transfer penalties.
Figure~\ref{fig:time_energy_cost_deconv_algos} shows how time and energy costs increase with upsampling factor.
When upsampling by a factor of 2, using REVD2 reduces latency by 1.6x and reduces energy per pixel by 1.9x \Note{in our experiments}.
As shown in Table~\ref{tbl:mem-requirements-algos}, the compute and activation requirements of transforming deconvolution to convolution (TDC) are the same as REVD2.
As such, the TDC zero-insertion requirements on the learned kernels have negligible impact on high resolution images.
However, as discussed in Section~\ref{sec:revd2}, the functional correctness breaks down when attempting to tile the workloads in sizes not evenly divisible by the stride $S$. 
While this penalty does not show in our simplified quantitative model of hardware performance, we aim to quantify this impact in future work.

% Algorithms with intensities such that $AI < B$ are ultimately memory-bound while algorithms with intensities such that $AI > B$ are ultimately compute-bound~\cite{williams2009roofline}.
% $B_\epsilon$ can then be interpreted as the point where memory operations no longer dominate the total energy cost~\cite{choi2013roofline}.
% As shown by Choi \textit{et al.}~\cite{choi2013roofline}, it is possible be 
% As shown in Figure~\ref{fig:resive-conv-activation-reuse}, even as $r$ increases, only D-SP is not dominated by memory operations.

% \Note{Note to self: We can only say this also implies time efficiency (i.e. higher throughput) when the energy-balance point is greater than the time-balance point for a given hardware architecture.}

\vspace{-0.3cm}
\section{Conclusions and Future Work}
\label{sec:conclusion}

Cloud computing systems can have nearly limitless resources, making them ideal for resource-intensive tasks such as data storage, data processing, and model training.
%However, edge computing paradigms are needed for real-time deep learning applications to decrease system latency by executing inference locally on edge devices without reliance on a stable network connection~\cite{dhar2019device, sze2017efficient}.
However, real-time deep learning applications often require edge computing frameworks to improve system latency by executing inference locally on edge devices without reliance on a stable internet connection~\cite{dhar2019device, sze2017efficient}.
We propose a novel edge computing paradigm for real-time convolution-based image upsampling applications that separately considers algorithms for training in the cloud and inference at the edge.
The use of sub-pixel or resize convolution is confined to training in the cloud to minimize the data transfer penalties incurred by the memory-intensive feature map transformations they require for inference.
The learned convolution kernels are then transformed to deconvolution kernels without sacrificing the image fidelity learned in training.
These deconvolution kernels are then deployed for inference at the edge using our improved reverse looping deconvolution algorithm (REVD2).
We compare REVD2 against pre-existing deconvolution variants and show it is more efficient and parallelizable.
Using quantitative models of time and energy, we show that executing deconvolution inference at the edge with REVD2 improves both system latency and energy efficiency when compared to sub-pixel or resize convolution counterparts.
When optimizing for energy efficiency and scalability, we show that training with sub-pixel convolution in the cloud and then transforming the learned kernels using the weight shuffle for deconvolution inference at the edge minimizes the pressure on memory bandwidth and maximizes energy efficiency.
When optimizing for latency and energy consumption, we show that training with nearest neighbor resize convolution in the cloud and then transforming the learned kernels using the weight convolution for deconvolution inference at the edge minimizes the time and energy costs incurred by upsampling an image.
In future work, we aim to extend our analyses to quantify adaptability to available hardware resources and build hardware designed to exploit the parallelism that is exposed from REVD2.
Code for each algorithm discussed in this paper can be found at \url{https://github.com/icolbert/upsampling}.

\pagebreak

\bibliography{citations.bib}

\end{document}